%% file: main.tex
\documentclass[manuscript,screen]{acmart}

\usepackage{amsmath}
\usepackage{booktabs}
\usepackage{graphicx}
\usepackage{float}
\usepackage{tikz}
\usetikzlibrary{arrows.meta,shapes.callouts,calc}
\usepackage{enumitem}
\usepackage{amsmath, amsthm}
\usepackage{longtable}
\usepackage{array}
\usepackage{enumitem}

\theoremstyle{acmplain}
\newtheorem{propos}{Proposition}
\newtheorem{corol}{Corollary}

\title{Personas Are Not People: Identity as State, Not Attribute}

\author{Tamunotonye Harry}
\affiliation{%
  \institution{University of Vermont}
  \department{Department of Computer Science}
  \city{Burlington}
  \state{Vermont}
  \country{USA}
}

\author{Yuanyuan Feng}
\affiliation{%
  \institution{University of Vermont}
  \department{Department of Computer Science}
  \city{Burlington}
  \state{Vermont}
  \country{USA}
}

\renewcommand{\shortauthors}{Harry and Feng}

\renewcommand\footnotetextcopyrightpermission[1]{}
\begin{document}

\begin{abstract}
\input{sections/00_abstract}
\end{abstract}


\maketitle

\begin{figure}[t]
  \centering
  \resizebox{0.90\linewidth}{!}{\input{figures/fig_intro_overview.tex}}
  \caption{(a) Today's simulated people (top) are written once, so the language model gets the same profile in every situation and any change in behavior is the model improvising in a way that fits anyone with that profile. In person-in-context simulation (bottom), the situation first brings out an identity state, such as parent, and the model acts out that state. (b) Our contributions are a proof that the top design cannot reproduce what is specific to a person (\S\ref{sec:falsification}), a criterion for measuring it (\S\ref{sec:evaluation-criterion}), and a test of one version of the bottom design on two real datasets (\S\ref{sec:empirical-tests}).}
  \label{fig:intro-overview}
\end{figure}
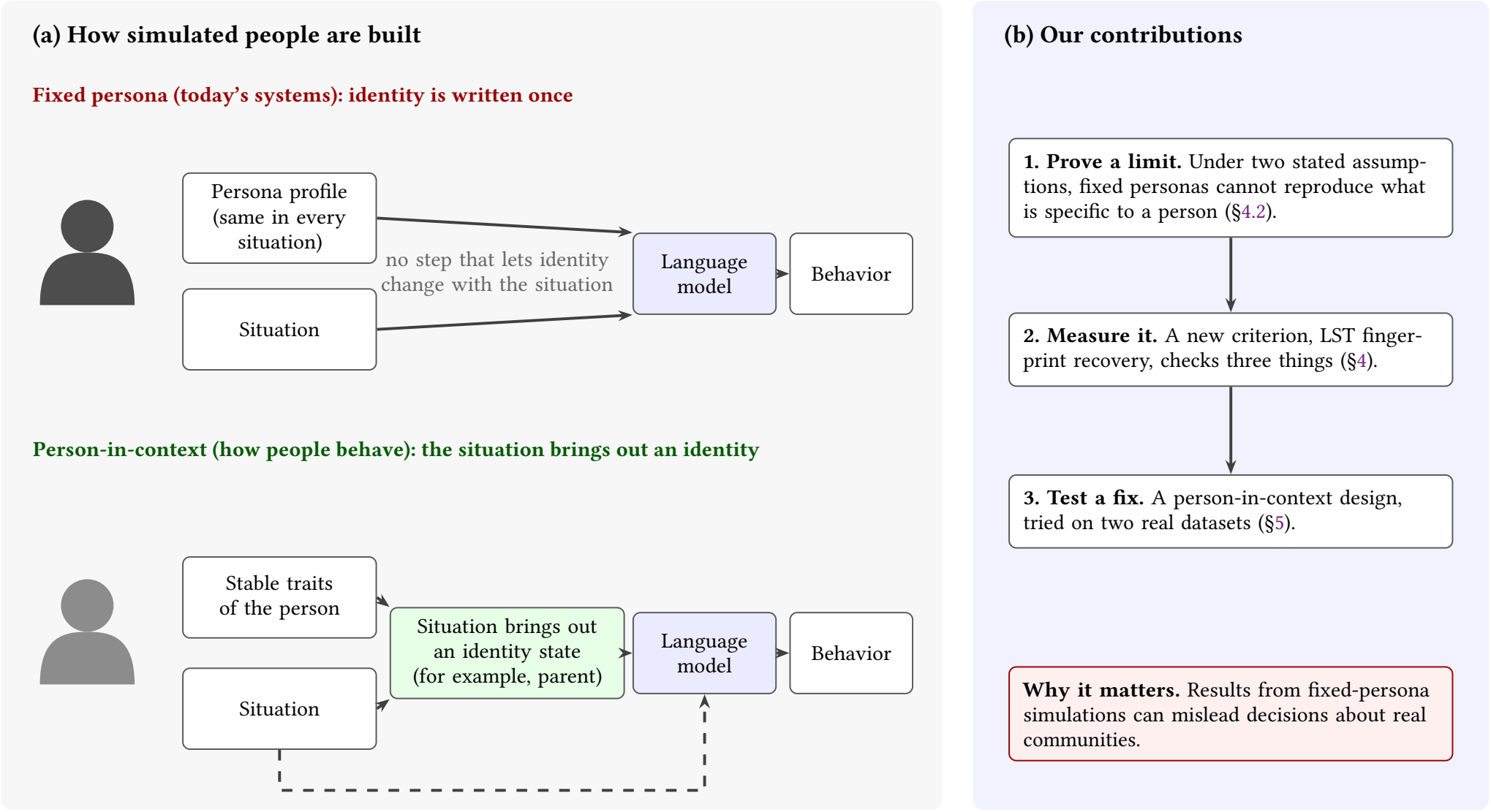

\input{sections/01_introduction}
\input{sections/02_related_work}
\input{sections/03_framework}
\input{sections/04_evaluation_criterion}
\input{sections/05_chameleon}

\input{sections/07_discussion}
\input{sections/08_endmatter}

\bibliographystyle{ACM-Reference-Format}
\bibliography{references}

\appendix
\input{sections/09_appendix}

\end{document}

%% file: sections/00_abstract.tex
Synthetic populations, large sets of simulated people, are increasingly used in place of real people to pretest surveys, simulate users and communities, and evaluate AI systems. Each simulated person is defined by a persona, a written profile of fixed traits, demographics, and attitudes that a language model is instructed to act out. Every such system assumes that a person's identity is a stable attribute that behaves the same way in every situation the simulation presents. We show that this assumption is not a harmless simplification. Three decades of personality research find the opposite. How a given person behaves varies more across contexts than people vary from one another. Drawing on identity economics, which describes how different situations activate different identities and norms within the same person, we separate behavior into three parts, which are stable traits, the identity a context calls up, and the situation itself. We then prove, under two stated assumptions, that any simulation built on fixed personas, no matter how detailed the personas or how large the underlying model, cannot reproduce the part of a population's behavior that is specific to it and its circumstances. We propose a new evaluation criterion based on this result and a person-in-context design that can meet it in principle. We test one version on a naturalistic Reddit personality corpus and on a published clinical-trial recruitment survey of a community with documented distrust of medical institutions. The design does produce context-dependent behavior when tested directly, but in neither setting does it fully match the real population, and we show where it falls short. The consequence for practice is a risk that synthetic-population evaluations misrepresent the groups whose behavior is most shaped by context, a distortion that richer personas cannot correct.

%% file: figures/fig_intro_overview.tex
\tikzset{pics/silhouette/.style={code={
  \fill[#1] (0,0.40) circle (0.20);
  \fill[#1] (-0.36,-0.20) .. controls (-0.36,0.17) and (-0.17,0.20) .. (0,0.20)
     .. controls (0.17,0.20) and (0.36,0.17) .. (0.36,-0.20) -- cycle;
}}}
\begin{tikzpicture}[x=1cm,y=1cm,font=\scriptsize,
  box/.style={draw=black!65,rounded corners=2pt,align=center,minimum height=0.8cm,inner sep=3pt,fill=white},
  hl/.style={draw=black!65,rounded corners=2pt,align=center,minimum height=0.8cm,inner sep=3pt,fill=green!10},
  lm/.style={draw=black!65,rounded corners=2pt,align=center,minimum height=0.8cm,inner sep=3pt,fill=blue!8},
  rbox/.style={draw=black!65,rounded corners=2pt,align=left,inner sep=4pt,fill=white,text width=4.1cm},
  arr/.style={-{Stealth[length=4pt]},thick,black!75},
  lab/.style={anchor=west,font=\scriptsize\bfseries}]
\fill[black!3,rounded corners=3pt] (-0.4,-3.7) rectangle (8.9,4.3);
\node[lab,font=\footnotesize\bfseries] at (-0.2,3.95) {(a) How simulated people are built};
\node[lab,text=red!60!black] at (-0.2,3.35) {Fixed persona (today's systems): identity is written once};
\pic[scale=1.3] at (0.45,1.55) {silhouette={black!70}};
\node[box,text width=1.7cm] (p)  at (2.35,2.15) {Persona profile\\(same in every\\situation)};
\node[box,text width=1.7cm] (s1) at (2.35,1.05) {Situation};
\node[lm,text width=1.2cm] (l1) at (6.55,1.6) {Language model};
\node[box,text width=1.0cm] (b1) at (8.0,1.6) {Behavior};
\draw[arr] (p.east)  -- (l1.north west);
\draw[arr] (s1.east) -- (l1.south west);
\draw[arr] (l1) -- (b1);
\node[align=center,text=black!60,text width=2.4cm] at (4.5,1.6) {no step that lets identity change with the situation};
\node[lab,text=green!35!black] at (-0.2,-0.15) {Person-in-context (how people behave): the situation brings out an identity};
\pic[scale=1.3] at (0.45,-2.2) {silhouette={black!45}};
\node[box,text width=1.7cm] (t)  at (2.35,-1.6) {Stable traits\\of the person};
\node[box,text width=1.7cm] (s2) at (2.35,-2.7) {Situation};
\node[hl,text width=2.1cm]  (a)  at (4.6,-2.15) {Situation brings out\\an identity state\\(for example, parent)};
\node[lm,text width=1.2cm]  (l2) at (6.55,-2.15) {Language model};
\node[box,text width=1.0cm] (b2) at (8.0,-2.15) {Behavior};
\draw[arr] (t.east)  -- (a.north west);
\draw[arr] (s2.east) -- (a.south west);
\draw[arr] (a) -- (l2);
\draw[arr] (l2) -- (b2);
\draw[arr,dashed] (s2.south) -- ++(0,-0.4) -| (l2.south);
\fill[blue!5,rounded corners=3pt] (9.2,-3.7) rectangle (14.3,4.3);
\node[lab,font=\footnotesize\bfseries] at (9.4,3.95) {(b) Our contributions};
\node[rbox] (r1) at (11.75,2.45) {\textbf{1. Prove a limit.} Under two stated assumptions, fixed personas cannot reproduce what is specific to a person (\S\ref{sec:falsification}).};
\node[rbox] (r2) at (11.75,0.85) {\textbf{2. Measure it.} A new criterion, LST fingerprint recovery, checks three things (\S\ref{sec:evaluation-criterion}).};
\node[rbox] (r3) at (11.75,-0.75) {\textbf{3. Test a fix.} A person-in-context design, tried on two real datasets (\S\ref{sec:empirical-tests}).};
\draw[arr] (r1.south) -- (r2.north);
\draw[arr] (r2.south) -- (r3.north);
\node[rbox,draw=red!60!black,fill=red!6] at (11.75,-2.75) {\textbf{Why it matters.} Results from fixed-persona simulations can mislead decisions about real communities.};
\end{tikzpicture}

%% file: sections/01_introduction.tex
\section{Introduction}
\label{sec:introduction}

Companies and researchers now pretest surveys, products, and messages on AI-simulated populations before asking real customers or citizens, and some propose using the simulations instead of asking at all \cite{argyle2023demographic,bisbee2024synthetic,li2026matraix}. Some systems already hold billions of simulated people \cite{li2026matraix}, built in part from hundreds of hours of real interviews \cite{park2024interviews}. These systems share one design choice. Each simulated person is a \emph{persona}, a written profile, vector, or interview transcript that is prepared before the simulation runs. Some systems retrieve different parts of it for different questions, but what is stored about the person never changes. A persona therefore fixes the person's \emph{identity} too, which we take from Akerlof and Kranton \cite{akerlofkranton2000} to mean a person's sense of self as a member of a social category, along with the norms for how members of that category act. A real person holds several identities, such as employee, parent, or friend, and which one is active depends on the situation, so the same person may be formal and careful in a meeting with colleagues and relaxed and joking at dinner with friends (\S\ref{sec:decomposition}). Current persona designs cannot represent this shift. Any change in behavior across situations is the large language model (LLM) improvising, and the persona itself never changes (Figure~\ref{fig:intro-overview}).

In personality psychology, most of the variation in trait-related behavior is within persons rather than between them \cite{fleeson2009implications,fleeson2001toward}. Identity economics \cite{akerlofkranton2000,akerlofkranton2010} offers an account of this variation. A person belongs to several social categories, each category carries norms for how its members act, and the situation determines which category is active and therefore which norms the person follows (\S\ref{sec:related-behavioral-science}). Recent work has started to notice the symptoms. Studies report that fixed personas collapse toward a typical default \cite{ozkan2026distribution} and compress variation within demographic groups \cite{wang2026lifemem}, while other work lets the personality of a single dialogue agent shift as a conversation unfolds \cite{pielage2026dynamic,wei2026situational}. To our knowledge, none of this work says what a simulated population must reproduce to behave like a real one, and none represents identity as a state that context activates (\S\ref{sec:related-work}). Existing validation does not fill that gap either. Persona adherence \cite{li2026matraix} asks whether a simulated person acts as its profile says. Test-retest consistency and prediction of held-out survey answers \cite{park2024interviews} are internal-validity checks that ask whether a simulation reproduces what a person answered on the same kind of survey. Subgroup coverage \cite{batzner2025representativeness} asks whether the right kinds of people are included. None of these asks the external-validity question of whether simulated people behave like real people once the situation changes. We call a test of that question a \emph{behavioral-fidelity evaluation} (do simulated people respond to a new situation the way real people do?).

The gap has consequences. Suppose a city emergency office pretests the wording of a flood evacuation alert on simulated residents, because a live test is impossible. A real resident may ignore a text from the city as one more automated message, yet leave at once when a neighbor knocks on the door or a relative calls. A persona that says ``distrusts authorities'' predicts that no version of the alert works, and a persona that leaves distrust out predicts that every version works. Either way, the office chooses its wording and its messenger on the model's guess, and the residents who most need to leave may stay. The error matters more when the simulated people stand in for patients or citizens and the decision reaches a real community (\S\ref{sec:discussion-representation}).

We make three contributions.

\begin{enumerate}[leftmargin=*]
  \item \textbf{\textit{Can richer personas fix this? No, and we give a formal proof.}} Any simulation built on a fixed persona cannot reproduce what is specific to a person beyond their group's typical response, however detailed the persona and however large the model. The proof relies on two stated assumptions, and we test how it degrades when they only roughly hold (\S\ref{sec:falsification}, Appendix~\ref{sec:appendix}).
  \item \textbf{\textit{How would we know? A new evaluation criterion.}} Latent State-Trait (LST) \emph{fingerprint recovery} has three parts. It asks how much of the variation in behavior is between people, how much comes from particular people in particular situations, and whether a simulation tracks each person's specific response. The proof shows that, in principle, fixed personas cannot meet the third check (\S\ref{sec:evaluation-criterion}).
  \item \textbf{\textit{What would work? A person-in-context design and two tests of it.}} We build a person-in-context approach, in which a situation selects an identity state that the language model then acts out, and we test it against fixed personas on two real datasets. Fixed personas make each person behave almost the same in every situation, far more than real people do, and our design comes closer on all five personality traits we tested. On a published clinical-trial recruitment study, our design reproduces the overall effect of trust on how respondents rate the ads, and its behavior changes as expected when the identity state is switched. It does not reproduce the study's central finding that which ad theme works best depends on how much a person trusts the healthcare system, and we explain why in \S\ref{sec:empirical-tests}. Person-specific signal is near zero for every system we test, so that test cannot yet separate them.
\end{enumerate}

In principle, a person-in-context design can represent person-specific response to context, which a fixed persona provably cannot. In our tests, the implementation we built reproduces some features of the real data, such as the effect of trust, but not yet the finding that the best ad theme depends on trust. Closing that gap is an open problem.

%% file: sections/02_related_work.tex
\section{Related Work}
\label{sec:related-work}

Our contribution connects three bodies of work that have not been brought together, namely synthetic populations for AI evaluation, behavioral science on how people act in context, and the critique of representational harm in generative systems.

\subsection{Synthetic populations for AI evaluation}
\label{sec:related-synthetic-pop}

Simulated people are used for LLM-based evaluation in many ways, for example to pretest survey questions, to simulate public opinion, and to stress-test chat systems with simulated users. These systems fall into three families, which differ in how they represent identity.

\emph{The fixed-persona family} dominates current systems. A persona, whether a categorical vector, a written profile, or an interview transcript, is set up in advance and does not change with the situation. MatrAIx \cite{li2026matraix} is the current large-scale example, with 8.3 billion persona records, a 1,290-dimension schema, and a 91.5\% persona-adherence rate. Park et al.\ \cite{park2024interviews} use two-hour interview transcripts as personas for 1,052 people, and report that, on held-out General Social Survey items, interview-based agents reach 83\% of participants' own test-retest consistency. Argyle et al.\ \cite{argyle2023demographic} condition on demographics with lighter personas. Bisbee et al.\ \cite{bisbee2024synthetic} find that such synthetic survey responses have much less variance than real respondents, and that many regression coefficients differ significantly from human data. In this family, context stays outside the persona, and the language model alone supplies any adaptation.

\emph{The memory-evolving family} adds accumulated experience to a fixed persona. Generative Agents \cite{park2023generativeagents}, SOTOPIA \cite{zhou2024sotopia}, Concordia \cite{vezhnevets2023concordia}, OASIS \cite{yang2024oasis}, and AgentSociety \cite{piao2025agentsociety} give agents a memory stream with reflection or planning. Variation then follows what has happened to the agent, closer to a diary than to identities that the present situation brings out.

\emph{The person-in-context family} treats identity as a state that context activates. It has theoretical roots in self-categorization theory and multiple-self-aspects frameworks \cite{mcconnell2011multiple}, and related computational work exists for single dialogue agents. Pielage et al.\ \cite{pielage2026dynamic} model an assistant's personality as latent states in a state machine whose transitions follow the conversation. Wei et al.\ \cite{wei2026situational} steer persona-related neurons in a model according to the situation. Lee et al.\ \cite{lee2025spectrum} build a persona from social identity, personal identity, and life context. These works adapt one agent within an interaction or enrich a single persona. We found none that represents a population of people with context-activated identities and tests it against real within-person variation. That is the gap we address.

\emph{Concurrent critiques.} Several recent studies conclude from different angles that fixed personas are the wrong unit for population simulation. They report collapse toward a typical default in some scenarios \cite{ozkan2026distribution}, declining observer-rated persona expression over long conversations \cite{gonnermann2026persona}, subgroup-specific distortions from multi-attribute personas \cite{morocho2026multiattribute}, inconsistent subgroup alignment from sociodemographic prompting \cite{sun2025sociodemographic}, and compressed variation within demographic groups, for which a longitudinal memory remedy is proposed \cite{wang2026lifemem}. None derives an evaluation criterion from within-person variance or proposes a context-activated identity design.

\subsection{Behavioral science on person-in-context}
\label{sec:related-behavioral-science}

In personality research, most of the variation in trait-related behavior is within persons, from one moment to the next \cite{fleeson2009implications}. This helped resolve the person-situation debate of the 1990s \cite{fleeson2001toward,fleesonjayawickreme2015}. Latent State-Trait (LST) theory \cite{steyer1992laten,steyer2015laten} splits behavior into trait, state, and error, and variance ratios akin to intraclass correlations give each part's share. Whole Trait Theory, which came later, describes a trait as a distribution of states around a stable level \cite{fleeson2001toward,fleesonjayawickreme2015}. Inspired by both, we write behavior as $Y = \tau + s(X) + \varepsilon$, a stable level plus a response to the situation plus noise. In \S\ref{sec:decomposition} we extend $s(X)$ with a person-specific, context-activated term $I_{p,c,t}$.

One study applies these methods to language-model behavior directly \cite{harry2026acl}. In a Reddit corpus, it finds that most variance is within persons, about 72 to 74\% across the 26 trait dimensions it measures, without removing measurement noise, in line with the psychological literature. It also finds that language models prompted with personas do not track context-driven change, which it calls \emph{state-blind}. Han et al.\ \cite{han2026personality} also use Whole Trait Theory to read LLM agents and find that one personality prompt is expressed differently across four conversational settings.

\emph{Identity as a building block.} Akerlof and Kranton \cite{akerlofkranton2000,akerlofkranton2010} treat identity as a variable in a person's decisions, with actions judged against the norms of the identity that is active. Trait Activation Theory \cite{tettguterman2000} reaches a compatible view, that situations activate traits through cues, and self-categorization theory \cite{turner1987selfcategorization} and the multiple-self-aspects framework \cite{mcconnell2011multiple} agree. We found no synthetic-population system or evaluation built on any of them.

\subsection{Representational harm in generative systems}
\label{sec:related-representational-harm}

Representational harm connects this work to questions of fairness and accountability. A synthetic population that flattens a group misrepresents that group, and decisions made from it reach real communities. Bender et al.\ \cite{bender2021parrots} warn that scaling language models without attention to whose language is represented risks amplifying dominant viewpoints. Blodgett et al.\ \cite{blodgett2020language} argue that work on ``bias'' often does not say which harms to which communities it addresses. Our criterion responds to this by naming one specific distortion, flattening of context-dependent behavior, and measuring it. Wang et al.\ \cite{wang2025roleplay} show that replacing human participants with LLMs can misportray and flatten identity groups. Two findings concern persona-based systems directly. Xiao et al.\ \cite{xiao2026collapse} find that models with the highest per-persona fidelity produce the most polarized populations, so persona adherence and population fidelity can pull apart. Batzner et al.\ \cite{batzner2025representativeness} find that only 35\% of 63 persona-based LLM studies report anything about representativeness. That is a separate problem, since a representative, transparently sampled set of personas is still fixed and would still fail our criterion. These critiques name the outcome but not its cause. We locate the flattening in one design choice, identity as a fixed attribute of the persona, which cannot match the real structure of within-person variation. Richer personas do not remove it.

\subsection{Gap}
\label{sec:related-gap}

Behavioral science models how people act in context, and identity economics accounts for the mechanism. Neither has been built into a synthetic-population system and tested there against real within-person variation, and current validation does not check for it. We add what the critique literature lacks (\S\ref{sec:related-synthetic-pop}, \S\ref{sec:related-representational-harm}), namely an explanation of why these failures occur, a new evaluation criterion derived from it (\S\ref{sec:evaluation-criterion}), and a formal result (\S\ref{sec:falsification}) that, under two stated assumptions, the failure comes from the design, whatever the scale or prompting.

%% file: sections/03_framework.tex
\section{Framework: Person-in-Context Simulation}
\label{sec:framework}

We split behavior into three parts, which are a stable attribute of the person, an identity state that the situation brings out, and the situation itself.

\subsection{A Decomposition of Behavior}
\label{sec:decomposition}

Let $p$ index persons, $c$ contexts, and $t$ the time point of an observation. A person can return to the same context at a different time point, for example the same subreddit a week later. In plain terms, behavior depends on who a person stably is, which identity the situation brings out, and what the situation demands. We write

\[
Y_{p,c,t} = f(\tau_p, I_{p,c,t}, X_{c,t}) + \varepsilon_{p,c,t}.
\]

\begin{itemize}
  \item \emph{$\tau_p$} is the person's stable attribute vector, made up of demographic attributes (such as age or income), habits, values, and dispositions that stay roughly the same across situations. Big Five inventories and values surveys measure the psychological part.
  \item \emph{$I_{p,c,t}$} is the identity state that the situation brings out. It is chosen from a set of possible identity states, each with its own norms for how a person in that state acts. Identity economics, Trait Activation Theory, self-categorization theory, and multiple-self-aspects frameworks motivate this (\S\ref{sec:related-behavioral-science}).
  \item \emph{$X_{c,t}$} is the situation, meaning the linguistic, social, and situational features of context $c$ at time $t$.
  \item \emph{$f$} produces behavior from the three inputs. In the systems we study it is a large language model, and our results do not depend on how it works inside.
  \item \emph{$\varepsilon_{p,c,t}$} is mean-zero noise specific to one time point, the part none of the other terms explain.
\end{itemize}

Applying the law of total variance twice, first conditioning on $\tau_p$ and then on $(I_{p,c,t},X_{c,t})$, splits the variance of $Y$ into a between-person part, a within-person part driven by $I$ and $X$, and noise:
\[
\mathrm{Var}(Y) = \underbrace{\mathrm{Var}_{\tau}[\mathbb{E}(Y\mid\tau)]}_{\text{between-person}} + \underbrace{\mathbb{E}_\tau[\mathrm{Var}(f\mid\tau)]}_{\text{within-person via }I,X} + \mathrm{Var}(\varepsilon).
\]

Variation within a person can arise even without $I$. A fixed persona still behaves differently when $X_{c,t}$ changes, because $f$ responds to the situation directly. What requires $I_{p,c,t}$ is narrower, namely the part of within-person variation that is \emph{specific to the person}. This is the interaction between who the person is and which situation they are in, which differs from the response to the situation that all persons share. Only when $I_{p,c,t}$ changes with context for a fixed person can this interaction be guaranteed to be person-specific. Otherwise it could be a side effect of how $f$ handles $(\tau_p,X_{c,t})$. This is the property that the criterion in \S\ref{sec:evaluation-criterion} detects. The decomposition has the standard form of multilevel models in personality psychology \cite{raudenbushbryk2002,snijdersbosker2011} and of Latent State-Trait theory \cite{steyer1992laten,steyer2015laten}. Our contribution is to use it as a specification for synthetic populations.

\subsection{Kinds of Systems and Their Interface}
\label{sec:taxonomy}

A simulation system stores a representation of each person, builds a prompt for each situation, and calls a language model. The inputs that the model's generation call receives are the system's \emph{interface}. A \emph{single-call} system produces each response with one such call. A system \emph{with subroutines} runs extra steps first, such as a rule or a second language-model call that scores which identity a situation brings out, and passes the result into the interface. The interface is then the full list of inputs to the generation call, whether they come from storage or from a subroutine. In the appendix, the generation function $f_{\mathrm{LLM}}$ also carries independent sampling noise $\eta_{p,c}$ (random variation in the generated text, so the same input can give different responses on different calls), and the time index is dropped, since the propositions concern the person-context decomposition only.

We call each design a \emph{family} and each implementation a \emph{system}. The families differ in which inputs reach the model, whatever the number of calls.

\begin{itemize}
  \item \emph{Fixed-persona family.} The system stores one persona $\pi_p$ that never changes, so the interface is $(\pi_p, X_{c,t})$ and only the situation changes with context. Any identity-like adaptation comes from the language model's own inference. MatrAIx \cite{li2026matraix}, Park et al.\ \cite{park2024interviews}, and Argyle et al.\ \cite{argyle2023demographic} are in this family.
  \item \emph{Memory-evolving family.} The system stores a persona and a growing record of past interactions $M_{p,t}$, so the interface is $(\pi_p, M_{p,t}, X_{c,t})$ \cite{park2023generativeagents,zhou2024sotopia,vezhnevets2023concordia,piao2025agentsociety}. The record grows with time and is not indexed by context. The record encodes the agent's history.
  \item \emph{Person-in-context family.} The system stores a stable attribute vector $\tau_p$ and a set $\mathcal{I}_p$ of identity states available to the person. A subroutine $\pi_I$ selects which state to activate, written $I_{p,c,t}\sim\pi_I(\tau_p,X_{c,t})$. The interface is $(\tau_p, I_{p,c,t}, X_{c,t})$. The set is fixed for each person, and only the realized state changes with context. This is the design we propose, and we know of no production system that uses it.
\end{itemize}

Adding subroutines to a fixed-persona system does not change its family if every subroutine reads only $\pi_p$ and $X_{c,t}$. Its output is then still a function of $(\pi_p,X_{c,t})$. What matters for the propositions is whether any interface input is both specific to the person and changes with context for that person. Only the person-in-context family has one.

Building $\mathcal{I}_p$ from data (\S\ref{sec:chameleon}, \S\ref{sec:jmir}) showed that the set can also include a \emph{baseline} state, a default that a person returns to when no context is strong, like the center of a person's state distribution in Whole Trait Theory. Whether the set has a baseline is an empirical question. The Chameleon set has one baseline and five activated states, and the motivational-themes set has no baseline and six activated states.

\subsection{A Limit Built Into the Fixed-Persona Design}
\label{sec:incapability}

We do not claim that fixed-persona and memory-evolving systems fail at what they were built for. Sampling varied profiles for product testing \cite{li2026matraix}, simulating social dynamics in a sandbox \cite{park2023generativeagents}, and predicting individual attitudes \cite{park2024interviews} are legitimate goals, judged fairly by their own criteria. Our claim is narrower. These systems are now proposed for \emph{behavioral simulation}, which means reproducing how a population responds across contexts. For that goal they meet a target their design does not represent.

In a fixed-persona system the persona does not change with context, so no input says which identity the situation has brought out for this person. The model must infer one from the persona and the situation, or a subroutine supplies one that reads only those two inputs. Either way the output is a function of $(\pi_p, X_{c,t})$ alone (Proposition 1, \S\ref{sec:falsification}), so person-context interaction is limited by the variance that the model's own inference adds \cite{han2026personality,harry2026acl}. This holds for every fixed-persona system, whatever the persona format or the model. A richer persona adds between-person variance and does not create person-context interaction. A similar argument applies to memory-evolving systems, whose memory is a within-person state but is not indexed by identity options the person can activate. Their variation follows the sequence of past events, with no role for the norms of the present situation.

The person-in-context family can represent person-specific response to context by construction. Whether a given implementation reproduces real within-person variance is an empirical question, which the criterion in \S\ref{sec:evaluation-criterion} answers. That the family is \emph{capable} of it is a property of the design, whatever the implementation. The fixed-persona failures in \S\ref{sec:chameleon} and \S\ref{sec:jmir} follow from this. The memory-evolving family is open to a similar argument, but neither dataset tests it, because both compare only fixed-persona arms (A and B) with the person-in-context arm (C). Arm C's results concern one implementation (our specification of the identity states and how they are activated). They do not change what the design itself can represent (\S\ref{sec:discussion-shows}).

%% file: sections/04_evaluation_criterion.tex
\section{Evaluation Criterion: LST Fingerprint Recovery}
\label{sec:evaluation-criterion}

\S\ref{sec:framework} shows that only the person-in-context family can represent $I_{p,c,t}$ as person-specific. A new criterion turns that difference into something measurable, which we call \emph{Latent State-Trait (LST) fingerprint recovery}. A fingerprint is a pattern specific enough to identify who it belongs to. A population's LST fingerprint is how stable people are relative to one another, and how much and in what way each person varies across situations. The criterion, built on the LST variance decomposition (\S\ref{sec:decomposition}) \cite{steyer1992laten,eiddiener2004}, checks whether a synthetic population reproduces the real fingerprint as well as the average response levels.

\subsection{The Criterion}
\label{sec:the-criterion}

LST fingerprint recovery compares a synthetic system $\mathcal{A}$ with a real population on three components.

\begin{enumerate}
  \item \emph{Between-person share (ICC).} The intraclass correlation (ICC, bias-corrected) is the fraction of all variation in behavior that comes from stable differences between people. The rest is one person varying across situations, plus noise. A system that inflates this share treats identity as more fixed than real behavior supports.
  \item \emph{Person-context interaction variance ($\mathrm{Var}_{P\times C}$).} This is the variance left after removing person and context main effects. It comes from particular people in particular contexts (\S\ref{sec:decomposition}). A system can match the real ICC and still show almost no interaction, because everyone reacts to context in the same way. ICC alone cannot tell these cases apart.
  \item \emph{Person-specific signal (person-specific residual correlation).} This is the component our proof addresses. A fixed persona cannot reach it by design, and a person-in-context system can in principle. Two people of the same general type, for example both skeptical of institutions, still react differently to a given situation. The shared, predictable part is \emph{state-typical}. The leftover part is \emph{person-specific}, the signature a system must capture if it claims to simulate that person, beyond their type.

  Formally, $R^{\text{human}}_{p,c} = m(\lambda_p,c) + \Delta_{p,c}$. Here $\lambda_p$ is person $p$'s soft membership across the identity states in $\mathcal{I}_p$ (\S\ref{sec:taxonomy}; Appendix~\ref{sec:appendix}), $m(\lambda_p,c)$ is the state-typical shift, and $\Delta_{p,c}$ is the person-specific residual. The third component is the correlation between the system's residual $R^{\mathcal{A}}_{p,c}$ and $\Delta_{p,c}$ only, leaving out the state-typical part $m(\lambda_p,c)$. The reason is that a system could match the full human residual $R^{\text{human}}_{p,c}$ by copying state-typical patterns that a stated persona often reveals, without representing anything specific to the person.
\end{enumerate}

A population that recovers all three, with real-scale ICC, real-scale interaction variance, and non-trivial person-specific signal, reproduces the target's LST fingerprint. A fixed persona falls short on the third component by design, as \S\ref{sec:falsification} proves. Whether a given person-in-context implementation reaches it is an empirical question.

\subsection{A Formal Limit for the Fixed-Persona Family}
\label{sec:falsification}

Under the two assumptions below, a fixed-persona system cannot meet the third component. This is a formal result about the design, so it does not depend on tuning or on any dataset. Full statements and proofs are in Appendix~\ref{sec:appendix}.

\emph{Mediation (Proposition 1; proof in Appendix~\ref{sec:appendix}).} The interaction residual of a fixed-persona system has the form $R^{\mathcal{A}}_{p,c} = g(\pi_p,X_c)+\eta_{p,c}$, where $\eta$ is independent sampling noise and $\mathrm{Cov}(g,\eta)=0$. It contains nothing beyond the persona, the context, and noise, however rich $\pi_p$ is. This limits the correlation with the person-specific residual. It does not limit how elaborate or context-sensitive $g(\pi_p,X_c)$ can be. No persona can make that sensitivity track $\Delta_{p,c}$, because $\pi_p$ does not change with context and carries no information indexed by person and context.

\emph{Person-specific non-transmission (Proposition 2; proof in Appendix~\ref{sec:appendix}).} A fixed-persona system's output cannot correlate with a real person's specific signature, and any apparent match comes from matching general type. The Key Assumption says that whoever builds the persona has no information about $\Delta_{p,c}$ beyond the person's identity state and demographics. The Context Exogeneity Assumption says that the context a person appears in also carries no such information. Under both, with the independence conditions in Appendix~\ref{sec:appendix}, $\mathrm{Cov}(R^{\mathcal{A}}_{p,c},\Delta_{p,c})=0$ exactly.

A robustness corollary relaxes both assumptions to approximate independence. If the persona leaks at most $\epsilon_1$ and the context assignment leaks at most $\epsilon_2$ of information about $\Delta_{p,c}$, then
\[
|\mathrm{Cov}(R^{\mathcal{A}}_{p,c},\Delta_{p,c})|\le 4M^2\sqrt{(\epsilon_1+\epsilon_2)/2}.
\]
Here $M$ bounds the size of the residuals, which is finite because ratings lie on a fixed scale. The quantity $\epsilon_1$ measures how much knowing the persona tells you about $\Delta_{p,c}$, beyond identity state and demographics, and $\epsilon_2$ measures the same for the context. Both are mutual information, and zero means no leak. The bound says a small leak allows only a small correlation, so the result degrades gradually and does not collapse. The Key Assumption is easier to defend for thin, schema-based personas than for rich ones, and the Context Exogeneity Assumption is easier to defend for experimentally assigned contexts than for observed ones.

\emph{What the proof adds beyond the definition.} The result would be empty if it only restated that a persona holds no information about $\Delta_{p,c}$. It says more. Proposition 1 shows that a fixed-persona output depends on the persona, the context, and noise only, for any language model and any persona, which rules out the common hope that richer personas or larger models will close the gap. The assumptions are not part of the definition, and they can fail, for example when a persona is built from a person's own behavior across contexts. The corollary bounds the effect when they hold only approximately. The prediction can also be tested. If personas leaked person-specific information, Arm A's correlation with the real residual would be large, and in \S\ref{sec:chameleon} it is small (0.026).

\emph{What this establishes.} No fixed-persona system that meets the two assumptions can reach the third component beyond a small margin set by the corollary, whatever the model scale or data. Richer prompts or larger models can improve the first two components without touching this limit.

\subsection{Scope: What the Criterion Does Not Establish}
\label{sec:criterion-scope}

The criterion has two limits. First, the proof concerns the third component only, so it does not say that fixed personas must show inflated ICC or near-zero interaction variance. Those are diagnostics that we report for every system in \S\ref{sec:chameleon}. A fixed-persona system could show real-scale ICC or substantial interaction variance and still fail the third component. Second, the result depends on the two assumptions, or their approximate version. They are claims about how personas are built and how contexts are assigned, and neither follows from context-invariance alone. A persona built from a demographic profile and a stated identity state meets the Key Assumption directly, because $\Delta_{p,c}$ is by construction what such instruments cannot see. A persona built by hand-picking a person's own responses across contexts would break it, because the persona would then contain the signal that Proposition 2 assumes it lacks.

Within this scope, the criterion gives person-in-context systems a target that fixed-persona systems cannot meet in principle, and gives any implementation of $\pi_I$ a concrete three-part measure. \S\ref{sec:chameleon} and \S\ref{sec:jmir} apply it to two datasets.

%% file: sections/05_chameleon.tex
\section{Empirical Tests of the Criterion}
\label{sec:empirical-tests}

The proof in \S\ref{sec:falsification} concerns what the fixed-persona design can represent. The experiments ask what the data show about fixed personas and what Arm C, a first, theory-motivated implementation of the alternative, achieves. In Chameleon, fixed personas (Arms A and B) place almost all variation between people, where real people show almost none, and in the motivational-themes study their agreement with the published pattern changes from 5 of 6 directions to 0 of 6 when one sentence is added. Arm C comes closest to the real variance structure on every Chameleon trait, it reproduces the effect of trust in the study, and its behavior follows the identity state it is given. Person-specific signal is near zero for every system we test, and Arm C does not yet reproduce the study's interaction between ad theme and trust. We trace that gap to how the implementation represents trust, as a single level that does not change with context (\S\ref{sec:jmir-results}), which is the flattening that \S\ref{sec:discussion-representation} names as a risk.

We chose the two datasets for what each can test. Chameleon records three posts from each of 1,667 people in different subreddits. Its repeated measures within each person, scored on many trait dimensions in a naturalistic corpus, make it the direct test of the between-person share and of person-specific signal. The motivational-themes study is not a dataset in the usual sense. It is a published survey experiment with one rating set per person, and we use its reported coefficients, sample demographics, and survey wording as a target. Because the real effect is known, we can say precisely where a simulation departs from it. Where the difference does not matter, we call both of them datasets.

\subsection{Experimental Setup}
\label{sec:shared-design}

Each dataset uses the same three arms, which differ only in what the language model is told about the person. Arm A (fixed persona) states a trait of the person once, such as ``skeptical of institutions'', whatever the context. Arm B (fixed persona plus context) states the same trait and adds a neutral mention of the context, which tests whether naming the situation is enough when the person does not change. Arm C (person-in-context) states no fixed trait. It computes an identity state for each (person, context) pair, and the model acts out that state. Arms A and B are fixed-persona systems and neither keeps a record of past interactions, so neither dataset tests the memory-evolving family (\S\ref{sec:incapability}). Appendix Table~\ref{tab:family-examples} shows what the model receives in each arm, and Appendix~\ref{sec:prompts-examples} gives the prompt for each arm.

Arm C builds its identity states in two steps. First, we construct a set of candidate identity states for the population. For Chameleon we fit a Gaussian mixture model (GMM), a statistical method that finds groups of similar observations in data without predefined labels. In our case the observations are users' trait scores, so the groups are users with similar trait profiles. For the motivational-themes study we write the states from clinical-recruitment literature, independently of the target study, and check them for circularity. Second, each person's affinity for each state is matched against the context to pick one active state. Figures~\ref{fig:chameleon-pipeline} and~\ref{fig:jmir-pipeline} in the appendix diagram the two pipelines. We fix each specification before testing it on real data. Any later revision is reported as a separate, documented version (Appendix~\ref{sec:regime-specs}).

All arms use the same elicitation format, in which the model gives a short reasoning and then a rating (Appendix~\ref{sec:elicitation-format-validation}), and the same model, Claude Haiku 4.5 at temperature 0.7. The model generates behavior in every arm. In Chameleon it also scores, in separate calls, how well each subreddit matches each identity state's cues. In the motivational-themes study, the cue scores for the four ads are fixed numbers set in the code by the authors from the specification's activation cues. We chose Claude Haiku 4.5 because its training data end in February 2025, before the motivational-themes study was published in March 2026, whereas later models may have been trained on its preprint and could recall the results.

The two datasets test different parts of the criterion. Chameleon tests the first component (between-person share) and the third, person-specific signal, directly. The motivational-themes study has one observation per person, so it cannot test the third component. We compare each arm with the published aggregate coefficients instead, which tests the second component, whether the system reproduces the real person-context interaction in aggregate.

\subsection{A Naturalistic Check: Chameleon}
\label{sec:chameleon}

This check asks whether each system reproduces a known pattern in real people, in which most of what varies in their behavior is within each person, from one situation to the next, and little is between people. The target is the Chameleon dataset \cite{harry2026acl}, with 1,667 Reddit users and three posts each in different subreddits. Each post is scored on twenty-six trait dimensions, including the five Big Five personality traits (extraversion, agreeableness, conscientiousness, neuroticism, and openness), which are the outcomes each simulation must reproduce. The other twenty-one dimensions, from values, self-determination, and risk-taking scales, describe the person in the persona. In the real data about 1\% of the variation is between people, and almost all of it is within each person across situations (\S\ref{sec:related-behavioral-science}). This uses a bias-corrected ICC, which removes the noise that three posts per person add.

On the first component, fixed personas show the opposite of the real pattern (Table~\ref{tab:chameleon-summary}). The real between-person share is about 1\% (mean ICC 0.013), and Arms A and B place it at 91\% and 80\%. Arm C is the closest to the real value on all five traits and on average (0.285), and it recovers extraversion almost exactly (0.023 against 0.019). On the other four traits it remains above the real value, by 24 to 65 times on agreeableness, conscientiousness, and openness, and below it on neuroticism. On the third component, person-specific signal is near zero for every arm (mean correlation 0.026 for Arm A, 0.025 for Arm B, and 0.004 for Arm C), and only extraversion is significant for all three (Table~\ref{tab:chameleon-summary}). This is the pattern Proposition 2 predicts for fixed personas. Their small positive values are what the robustness corollary (\S\ref{sec:falsification}) expects, since a persona built from scores of the user's own posts, and subreddits that users chose for themselves, can carry a little information about that user's specific responses. With about 4,570 comparisons per trait, the test can reliably detect a correlation of about 0.04 or more (80\% power, Appendix~\ref{sec:chameleon-full-test}). It therefore rules out a correlation well above that size in every arm and cannot separate the arms below it, so whether the person-in-context design recovers person-specific signal remains open.

\begin{table}[t]
  \centering
  \small
  \caption{Chameleon results by trait and arm. ICC is the bias-corrected between-person share, and bold marks the arm closest to the real value. Person-specific correlation is the correlation with the real residual $\Delta_{p,c}$, bold marks the highest value in a row, and an asterisk marks a 95\% interval that excludes zero (Appendix Tables~\ref{tab:chameleon-icc} and~\ref{tab:chameleon-idio}).}
  \label{tab:chameleon-summary}
  \begin{tabular}{l@{\hspace{8pt}}rrrr@{\hspace{10pt}}rrr}
    \toprule
     & \multicolumn{4}{c}{ICC (between-person share)} & \multicolumn{3}{c}{Person-specific correlation} \\
    \cmidrule(r{6pt}){2-5}\cmidrule(l){6-8}
    Trait & Real & Arm A & Arm B & Arm C & Arm A & Arm B & Arm C \\
    \midrule
    Extraversion      & 0.019    & 0.954 & 0.861 & \textbf{0.023}    & $0.039^{*}$ & $\mathbf{0.064}^{*}$ & $0.030^{*}$ \\
    Agreeableness     & 0.009    & 0.933 & 0.891 & \textbf{0.574}    & 0.010       & \textbf{0.012}       & $-$0.020 \\
    Conscientiousness & 0.022    & 0.901 & 0.800 & \textbf{0.585}    & $\mathbf{0.033}^{*}$ & 0.012       & $-$0.005 \\
    Neuroticism       & $-$0.007 & 0.828 & 0.596 & \textbf{$-$0.248} & $\mathbf{0.035}^{*}$ & $0.028^{*}$ & 0.004 \\
    Openness          & 0.020    & 0.930 & 0.846 & \textbf{0.489}    & \textbf{0.011} & 0.008       & \textbf{0.011} \\
    \midrule
    Mean              & 0.013    & 0.909 & 0.799 & \textbf{0.285}    & \textbf{0.026} & 0.025       & 0.004 \\
    \bottomrule
  \end{tabular}
\end{table}

\subsection{A Published Target: The Motivational-Themes Study}
\label{sec:jmir}

The motivational-themes study \cite{harry2026jmir} asked which advertising themes make a clinical-trial recruitment ad effective for Black adults in the United States, measured by how appealing respondents found the ad and how willing they were to participate and to recommend the trial. It also measured each respondent's trust in the healthcare system, because distrust of medical institutions is documented in this community \cite{scharff2010more,armstrong2013prior} and could change how a theme lands. We use its published results as the target.

\subsubsection{Setting and Reproduction Target}
\label{sec:jmir-setting}

The study surveyed 829 Black adults. Each saw one of four ad themes (contribution to science, community benefit, blood-pressure management, financial compensation), a between-subject design. They rated appeal, willingness to participate, and willingness to recommend on 5-point scales. The target has two parts. First, no theme has a significant main effect on any outcome ($p=.25$ to $.93$). Second, theme and trust interact for the blood-pressure theme. All six blood-pressure by trust odds ratios are below 1 at \emph{both} trust extremes relative to neutral, which means lower odds of a favorable rating. Significance alternates by outcome between the trust side and the distrust side. The target is this six-coefficient directional pattern, which is stricter than significance at both extremes (Appendix Table~\ref{tab:jmir-target} lists the published design and values). For each arm we estimate the same six interactions and compare their signs with the published ones. The main analysis is a linear model in rating points with the contribution-to-science theme and neutral trust as baselines, and Appendix Table~\ref{tab:jmir-specs} repeats the comparison under seven analysis specifications.

The theme main effect does not separate the families, but the interaction is informative, since it requires the \emph{same} theme to evoke a \emph{different} response depending on a person-level trait. A fixed persona can reproduce part of this pattern in aggregate, because the language model links a stated trait to the ad text on its own. It cannot represent an identity that changes with the theme. The aggregate interaction is therefore a coarse test of whether a system represents the shift, since it cannot show whether any single simulated person changed.

\subsubsection{Materials and Cohort}
\label{sec:jmir-harness}

All three arms see the study's advertisement text and outcome wording word for word. Arm C uses six identity states written from the literature (Appendix~\ref{sec:regime-spec-jmir}). The Medical Skeptic state, on which the target interaction depends, is driven mainly by trust tier (activation probability 0.70 for distrust, 0.35 for neutral, 0.15 for trust), with capped demographic modifiers that cannot outvote trust. Appendix~\ref{sec:regime-spec-jmir} gives the calibration history.

Assignment is between-subject, as in the original study. We use 6,632 demographically resampled personas, one theme each, and all three arms run on identical cells. As \S\ref{sec:shared-design} explains, Claude Haiku 4.5's training data end in February 2025, thirteen months before the study's publication, which makes contamination unlikely.

\subsubsection{Results}
\label{sec:jmir-results}

\textbf{\emph{The generative step responds to the active state.}} In the full run, 479 of 6,632 personas received, by chance, at least one response under an active Medical Skeptic state and one under their theme's own state. The paired within-person difference is large and consistent ($-1.76$ appeal, $-1.24$ participate, $-2.44$ recommend), with the same direction in 87 to 100\% of personas (Appendix Figure~\ref{fig:within-person-slopes}). A forced, matched 30-per-condition check gives non-overlapping 95\% intervals. Switching a persona's active state reliably changes behavior in the expected direction.

\textbf{\emph{At $n=6{,}632$, Arm C reproduces the trust main effect but not the interaction.}} An initial $n=829$ run, matching the original study, was ambiguous. Interaction effects need larger samples than main effects to detect, so the cohort grew to $n=6{,}632$, with roughly 500 to 630 personas per critical cell. At this size Arm C shows a large, precise trust main effect. Mean appeal rises from 2.41 (distrust) to 3.04 (neutral) to 3.42 (trust), tracking the activation rate of the Medical Skeptic state (0.78, 0.43, 0.22; cell means in Appendix Table~\ref{tab:jmir-cellmeans}). For the six blood-pressure by trust coefficients, we compare the direction of each simulated interaction with the published one (Appendix Table~\ref{tab:jmir-specs}, row 2). Arm C matches 2 of 6, and both matches are coefficients close to zero. Arm A matches 5 of 6. Arm B, which adds only a theme sentence to Arm A, matches 0 of 6. Two arms that differ by one sentence give opposite directions, so a fixed persona's agreement with the published direction reflects incidental wording, so it is weak evidence that the persona represents the mechanism. Guessing would match 3 of 6 on average, so a count is a coarse test. The larger sample ruled out under-sampling as the explanation, and did not produce the interaction.

\textbf{\emph{The diagnosis concerns structure.}} The Medical Skeptic state's activation floor is the same for all four themes by design. It can shift the \emph{level} of activation by theme, but not the \emph{slope} of the response to trust. Only a theme-dependent slope can produce an interaction. Without one, the model yields a general main effect, and this limit holds at any sample size. Arm C also rates the blood-pressure theme lower than the contribution-to-science theme at every trust level (mean appeal 2.78 against 3.16 at neutral trust; Appendix Table~\ref{tab:jmir-cellmeans}), a theme main effect the real study did not find. For Arm A the limit comes from the design (Propositions 1 and 2). For Arm C it is a gap in this particular $\pi_I$, and the natural-experiment check shows the mechanism itself is not barred from closing it (Appendix Table~\ref{tab:cross-domain-synthesis} summarizes both datasets).

\begin{figure}[t]
  \centering
  \includegraphics[width=0.72\linewidth]{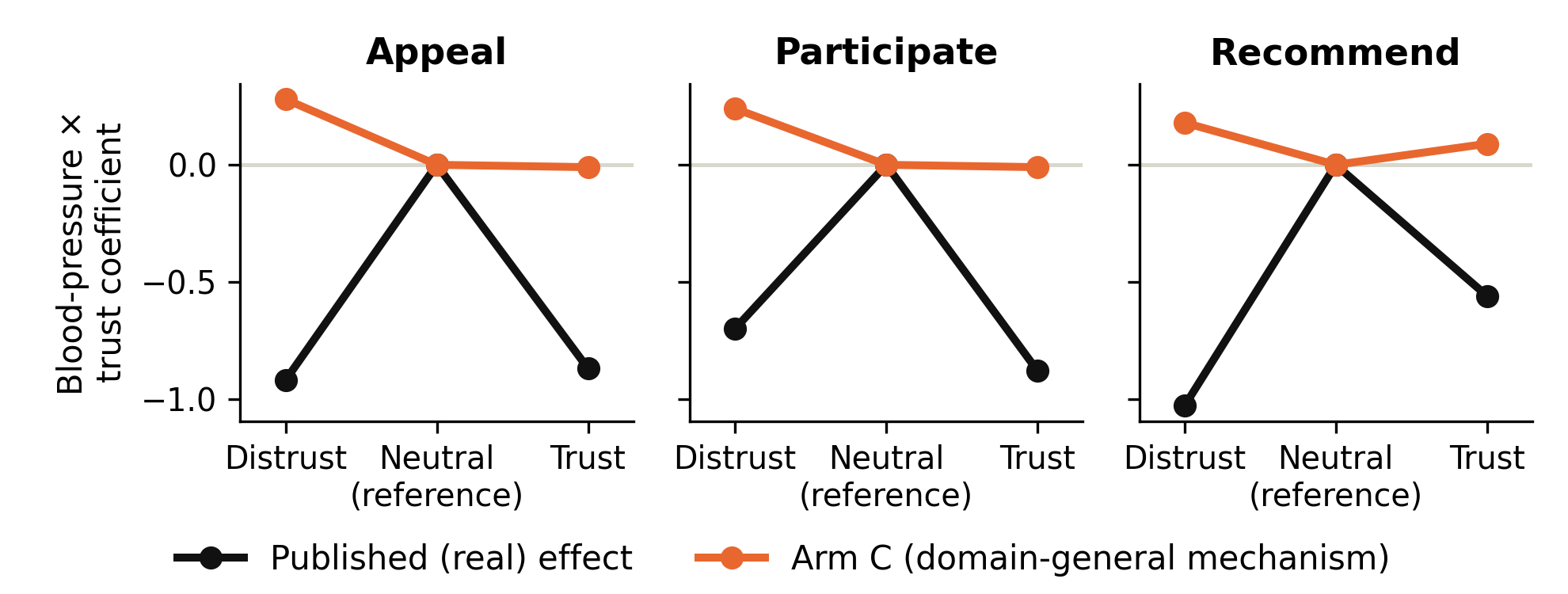}
  \caption{The real published effect dips at \emph{both} trust extremes (all six coefficients negative), while the domain-general mechanism shows no dip at either extreme (all six simulated coefficients are $-0.01$ or higher). Published coefficients are log-odds and simulated ones are rating points, so the figure compares shape and sign only. A single-axis activation mechanism cannot resolve this mismatch by calibration alone.}
  \label{fig:shape-mismatch}
\end{figure}

\textbf{\emph{A theory-motivated refinement does not close the gap.}} Amplifying the trust-linked state's activation spread for the blood-pressure theme (fixed before evaluation, Appendix~\ref{sec:regime-spec-jmir}) does not raise the match count under the contribution-to-science baseline (1 of 6, against 2 for the domain-general mechanism). With a pooled-theme baseline it reaches 3 of 6, and Arm A falls to 2 or 3, so the order of these arms depends on the baseline. We make no claim that the refinement improves on the domain-general mechanism, and because the amplified theme was chosen knowing where the real effect lies, it is also a weaker test. The more useful result is the shape of the real effect. All six published coefficients are negative. High-trust and high-distrust respondents both rate the blood-pressure theme lower than neutral-trust respondents. A mechanism that moves along one trust axis cannot produce a dip at both extremes (Figure~\ref{fig:shape-mismatch}).

\subsection{Interpretation: What Both Datasets Show}
\label{sec:empirical-interpretation}

Neither dataset tests whether the person-in-context family is \emph{capable} of context-dependent behavior, since the natural-experiment and forced-contrast checks in \S\ref{sec:jmir-results} show that it is. They test whether this implementation of $\pi_I$ closes the gap in practice, which Proposition 2 does not address.

The failure can be diagnosed. The real effect dips at both trust extremes, a shape that no single-axis mechanism can represent (Figure~\ref{fig:shape-mismatch}), which is why the theme-amplified refinement produces no gain in Appendix Table~\ref{tab:jmir-specs}. Chameleon adds a complementary view. Its real within-person signal is faint, so it tests whether the implementation's gain in aggregate variance carries over to person-specific signal, and so far it does not (\S\ref{sec:chameleon}). Appendix Table~\ref{tab:cross-domain-synthesis} sets the two datasets side by side. Together they show what the implementation already does, which is to produce behavior that follows the active state, and where work remains, in the choice of state and its link to person-specific signal.

%% file: sections/07_discussion.tex
\section{Discussion}
\label{sec:discussion}

\subsection{What the empirical tests show, and what they do not}
\label{sec:discussion-shows}

The proof and the implementation are separate questions. Propositions 1 and 2 (\S\ref{sec:evaluation-criterion}) concern the design, and the experiments concern one implementation of the alternative. The difficulty in those tests concentrates in $\pi_I$, the step that selects the identity state, while the generative step, $f$, performs well (\S\ref{sec:empirical-interpretation}).

Some of what we observe comes from the language model as well as from the design. Ratings follow a stated trait almost mechanically, and Arm A's appeal ratings sit close to 2, 3, and 4 for distrust, neutral, and trust in every theme (Appendix~\ref{sec:jmir-robustness}). One added sentence reversed five of the six interaction coefficients (Arm B). And the model responds strongly and in the expected direction when its active identity state is switched (\S\ref{sec:jmir-results}), so the weak point is choosing the state. Acting it out works. The theory does not supply these facts about language models as simulators, and they should be rechecked with other models.

\subsection{What this changes for evaluation practice}
\label{sec:discussion-evaluation}

Persona adherence is the right check for the tasks that fixed-persona and memory-evolving systems were built for, namely covering varied profiles \cite{li2026matraix}, predicting individual attitudes \cite{park2024interviews}, and simulating sandbox dynamics \cite{park2023generativeagents}. It is the wrong check when such systems are proposed for behavioral simulation, because \S\ref{sec:incapability} shows that a design can satisfy persona adherence and still be unable to produce the person-context interaction that behavioral simulation needs. Any claim of behavioral simulation should report the criterion in \S\ref{sec:evaluation-criterion} directly instead of inferring it from adherence or demographic accuracy.

The motivational-themes study adds guidance for the one family that can simulate behavior in principle. Because the failure sits in $\pi_I$, auditing the generative step alone checks necessary behavior but not sufficient behavior, so evaluations should also audit $\pi_I$ against real population data (Appendix~\ref{sec:regime-spec-jmir}) and should not treat a literature-grounded specification as self-validating.

\subsection{What this changes for identity representation}
\label{sec:discussion-representation}

The fixed-persona design misrepresents groups whose behavior depends on context, and richer personas do not remove the problem. The distrust example from the introduction applies here, since a fixed persona stores distrust as a static attribute or leaves it out, and neither represents a state that the situation activates, as with distrust of medical institutions in the motivational-themes study \cite{scharff2010more,armstrong2013prior}. \S\ref{sec:jmir-results} shows the risk is concrete, because our own person-in-context design, built to represent trust as context-dependent, produces a version of the same flattening. The risk therefore appears whenever a mechanism that selects identity states under-represents how a trait depends on context.

This also moves an accountability question without removing it. A fixed persona's content can be traced to a stated source, such as demographic conditioning or a profile instrument. A person-in-context system also needs someone to specify $\mathcal{I}_p$ itself (the set of identity states available to each person, \S\ref{sec:taxonomy}), meaning which identity states exist, what activates them, and whether a behavior counts as state-typical or person-specific. That choice is an interpretation. Our own mechanism needed four locked revisions (documented versions, each fixed in writing before it was run and each prompted by a failed internal check) before it passed those checks, and the first three versions, though literature-grounded, failed them. The question shifts from what the persona says to who decides what a person's possible identity states are.

\subsection{Limitations}
\label{sec:discussion-limitations}

\emph{Data.} The 6,632 simulated personas in the motivational-themes study match the published share of each demographic group, such as age or income, but not how those characteristics combine in real respondents. The Chameleon test uses the real users, with three posts each (\S\ref{sec:chameleon}). \emph{Method.} The identity states are written by hand, so the results concern this implementation and say nothing about learned mechanisms (\S\ref{sec:discussion-forward}). In Chameleon, Claude Haiku 4.5 also scores how well each subreddit matches each identity state (Appendix~\ref{sec:regime-specs}), in a separate call from the one that generates behavior, so shared biases may carry through, a concern raised for LLM-as-judge setups \cite{li2026preferenceleakage}, and we did not test for this. In the motivational-themes study the cue scores were set by the authors, and the Medical Skeptic score was revised after a pilot (Appendix~\ref{sec:regime-spec-jmir}). The three identity-state panels were isolated agent sessions of Claude Sonnet 5.5, one model family with no human raters. Their agreement rules out one unlucky sample, but not a bias that all three share. \emph{Scope.} The framework models cross-sectional response only, and the between-subject motivational-themes study limits that test to the aggregate interaction. The proof does not depend on any of these.

\subsection{Forward}
\label{sec:discussion-forward}

Three directions follow. First, \S\ref{sec:jmir-results} motivates activation mechanisms that do not move along a single trust axis, since the real effect dips at both trust extremes. One outcome can arise from different states, for example institutional distrust at one extreme and a different skepticism at the other. Second, it is open whether the baseline-plus-activated-states structure in \S\ref{sec:taxonomy} is general or specific to Chameleon. A related test is whether Arm C does best on the trait whose context cues are clearest, which could explain why it recovers extraversion (\S\ref{sec:chameleon}). We have not tested this. Third, because the difficulty sits in $\pi_I$, it could be learned against held-out real data instead of set by hand \cite{wu2026humanlm}, for example with an input-driven switching state-space model in which context drives transitions between identity states and a language model writes the behavior from the current state. Testing it needs longitudinal data that our datasets lack.

%% file: sections/08_endmatter.tex
\section*{Ethical Considerations}
\label{sec:endmatter}

\emph{Data and human subjects.} We collected no new human-subjects data, and both datasets are already published. Chameleon is a public dataset of pseudonymous trait scores computed from public Reddit posts. It keeps the original post identifiers, so a record could in principle be linked to a public post. We made no attempt to identify, link, or contact any user, and we report results only in aggregate. The motivational-themes study contributes its published coefficients, sample demographics, and survey wording \cite{harry2026jmir}. We used the advertisement text only as a fixed input to language models, and did not use it to recruit or contact anyone. An institutional review board determined through its self-determination process that the project is not human-subjects research.

\emph{Population construction and the risk of encoding stereotypes.} The identity-state specification in \S\ref{sec:jmir} was built for a population, Black adults evaluating clinical-trial recruitment materials, for whom misrepresentation has real stakes, given the documented history of harm and eroded trust around clinical research \cite{scharff2010more,alsan2018tuskegee}. Three steps address this. The state content and demographic modifiers came from the independent health-behavior and clinical-recruitment literature, without using the target study's results (Appendix~\ref{sec:regime-spec-jmir}). Three isolated, blind sessions of Claude Sonnet 5.5 checked the specification without access to the target study's findings (three sessions of one model, which are not independent raters; see \S\ref{sec:discussion-limitations}). And the paper's central claim is that institutional distrust in this population is \emph{context-activated}, so assigning a static ``distrusts the healthcare system'' trait to a demographic group is the essentializing move the paper argues against. This does not remove the risk of misuse, because a published state specification could be separated from its context-activation framing.

\emph{Deployment risk.} \emph{This work does not support replacing engagement with underrepresented populations by simulation, and our results are evidence against doing so.} A reader could conclude that a validated person-in-context system could substitute for focus groups, surveys, or community engagement, especially for populations that are costly to recruit. That is where substitution would do the most harm. Our mechanism was literature-grounded, checked by blind model sessions, and run on a sample eight times the size of the original study, and it still did not reproduce the published interaction. This weakens the case for treating current synthetic output as a substitute for the represented population. These methods suit stress-testing evaluation infrastructure, but decisions about real populations need independent validation.

\emph{Reproducibility.} We document the specifications, the main calibration decisions, and the fixed parameters behind each result (Appendices~\ref{sec:chameleon-full-test}, \ref{sec:regime-spec-chameleon}, and \ref{sec:regime-spec-jmir}). Code, data, and saved model outputs are at \url{https://anonymous.4open.science/r/state-persona-eval/}.

\emph{Generative AI usage statement.} Generative AI tools were used for editing assistance and for re-running analyses to check reported numbers against the raw output of each experiment. The authors designed the study, ran the experiments, and take full responsibility for the content of this paper, including the accuracy of all claims, numbers, and citations. Language models are also the subject of the study, as the systems under evaluation.

%% file: sections/09_appendix.tex
\section{Extended Notation}
\label{sec:notation}

This section lists the symbols used in the main text and in the proofs. The proofs drop the time index $t$, since they concern the person-context decomposition only.

\begin{center}
\renewcommand{\arraystretch}{1.25}
\small
\begin{longtable}{@{}p{2.3cm}p{9.4cm}p{2.5cm}@{}}
\toprule
\textbf{Symbol} & \textbf{Meaning} & \textbf{Defined in} \\
\midrule
\endhead
\multicolumn{3}{@{}l}{\emph{Indices and behavior}} \\
$p$, $c$, $t$ & A person, a context, and the time point of an observation. & \S\ref{sec:decomposition} \\
$p'$, $c'$ & A person or context other than $p$ or $c$, used in the proofs. & Prop.~\ref{prop:non-transmission} \\
$Y_{p,c,t}$ & Real behavior (for example a rating) of person $p$ in context $c$ at time $t$. The proofs write $Y_{p,c}$, or $Y^{\mathrm{human}}_{p,c}$ when it must be told apart from simulated behavior. & \S\ref{sec:decomposition} \\
$\hat Y^{\mathcal{A}}_{p,c}$ & Behavior generated for $p$ in $c$ by a design $\mathcal{A}$, meaning a class of systems such as the fixed-persona family. & Def.~\ref{def:fixed-persona} \\
\midrule
\multicolumn{3}{@{}l}{\emph{The decomposition}} \\
$\tau_p$ & Stable attributes of the person: demographics, habits, values, and dispositions. & \S\ref{sec:decomposition} \\
$I_{p,c,t}$ & The identity state that the situation brings out for $p$. & \S\ref{sec:decomposition} \\
$\mathcal{I}_p$ & The set of identity states available to person $p$. & \S\ref{sec:taxonomy} \\
$\pi_I$ & The rule that selects the active identity state from $\tau_p$ and the situation. & \S\ref{sec:taxonomy} \\
$X_{c,t}$, $X_c$ & The situation, meaning the context input such as the text of an ad or the name of a subreddit. & \S\ref{sec:decomposition} \\
$f$ & The function that produces behavior from $\tau_p$, $I_{p,c,t}$, and $X_{c,t}$. & \S\ref{sec:decomposition} \\
$\varepsilon_{p,c,t}$ & Noise specific to one time point. & \S\ref{sec:decomposition} \\
\midrule
\multicolumn{3}{@{}l}{\emph{Systems}} \\
$\pi_p$ & The persona of person $p$ in a fixed-persona system: a profile, vector, or transcript that does not change across contexts. & Def.~\ref{def:fixed-persona} \\
$M_{p,t}$ & The record of past interactions in a memory-evolving system. & \S\ref{sec:taxonomy} \\
$f_{\mathrm{LLM}}$ & The generation function, a language model conditioned on the interface inputs. & Def.~\ref{def:fixed-persona} \\
$\eta_{p,c}$ & Sampling noise: random variation in generated text when $\pi_p$ and $X_c$ are held fixed. It is independent of both. & Def.~\ref{def:fixed-persona} \\
\midrule
\multicolumn{3}{@{}l}{\emph{The criterion}} \\
ICC & Intraclass correlation, the share of variance that lies between persons. We use the bias-corrected form. & \S\ref{sec:the-criterion} \\
$\mathrm{Var}_{P\times C}$ & Person-context interaction variance. & \S\ref{sec:the-criterion} \\
$Z_{p,c}$ & Any person-by-context variable, used to define the residual below. $\bar Z_{p,\cdot}$, $\bar Z_{\cdot,c}$, $\bar Z_{\cdot,\cdot}$ are its averages over contexts, over persons, and over both. & Def.~\ref{def:interaction-residual} \\
$R_{p,c}(Z)$ & The interaction residual: $Z_{p,c}$ minus the person average and the context average, plus the grand average. & Def.~\ref{def:interaction-residual} \\
$R^{\mathcal{A}}_{p,c}$, $R^{\mathrm{human}}_{p,c}$ & The interaction residual of a design's simulated behavior and of real behavior. & Def.~\ref{def:interaction-residual} \\
$\lambda_p$, $k$ & Soft membership of $p$ across $k$ identity-state components (weights that sum to 1). & Def.~\ref{def:regime-decomposition} \\
$S_p$ & A discrete identity-state label, the case where all weight of $\lambda_p$ is on one component. & Def.~\ref{def:regime-decomposition} \\
$D_p$ & Demographics of person $p$. & Key Assumption \\
$m(\lambda_p,c)$ & The state-typical shift: the part of $R^{\mathrm{human}}_{p,c}$ predictable from $\lambda_p$ and $c$. & Def.~\ref{def:regime-decomposition} \\
$\Delta_{p,c}$ & The person-specific deviation: what remains of $R^{\mathrm{human}}_{p,c}$ after removing $m(\lambda_p,c)$. & Def.~\ref{def:regime-decomposition} \\
\midrule
\multicolumn{3}{@{}l}{\emph{The proofs}} \\
$\mu(\pi_p)$, $\nu(X_c)$, $\bar\mu$ & Average simulated behavior of persona $\pi_p$ over contexts, of context $X_c$ over personas, and of both. & Prop.~\ref{prop:boundedness} \\
$g(\pi_p,X_c)$ & The part of $R^{\mathcal{A}}_{p,c}$ that depends only on persona and context, so $R^{\mathcal{A}}_{p,c}=g+\eta_{p,c}$. & Prop.~\ref{prop:boundedness} \\
$I(\cdot\,;\cdot\mid\cdot)$ & Conditional mutual information, zero exactly when two variables are conditionally independent. & Cor.~\ref{cor:robustness-single} \\
$\epsilon$, $\epsilon_1$, $\epsilon_2$ & Bounds on how much the persona ($\epsilon_1$), the context ($\epsilon_2$), or both together ($\epsilon=\epsilon_1+\epsilon_2$) reveal about $\Delta_{p,c}$. & Cor.~\ref{cor:robustness-joint} \\
$M$ & A bound on $|R^{\mathcal{A}}_{p,c}|$ and $|\Delta_{p,c}|$, finite because ratings lie on a fixed scale. & Cor.~\ref{cor:robustness-single} \\
$\perp$ & Statistical independence. $A\perp B\mid C$ means that given $C$, $A$ carries no information about $B$. & Key Assumption \\
$\mathrm{Cov}(\cdot,\cdot)$ & Covariance. & Prop.~\ref{prop:boundedness} \\
\bottomrule
\end{longtable}
\end{center}

\section{Fixed-Persona Design-Level Limit: Core Propositions}
\label{sec:appendix}

\noindent\emph{Setting.} Notation is in Appendix~\ref{sec:notation}. Proposition~\ref{prop:boundedness} treats the averages in Definition~\ref{def:interaction-residual} as population expectations. With finite samples they add terms from other persons, and Proposition~\ref{prop:non-transmission} handles these through independent sampling of persons, so that the persona, context, and noise terms of other persons are independent of person $p$'s own $\Delta_{p,c}$. The sampling noise $\eta_{p,c}$ is independent of all persona, context, and human quantities. Proposition~\ref{prop:non-transmission} also uses one further condition, that given $(\lambda_p,D_p,\Delta_{p,\cdot})$ the persona and the context are independent. It holds when the persona is built without reference to the contexts later used. With the two named assumptions it gives joint independence of $(\pi_p,X_c)$ and $\Delta_{p,\cdot}$ (Step 1).

\subsection{Definitions}

\begin{definition}[Fixed-persona design]\label{def:fixed-persona}
A design $\mathcal{A}$ is \emph{fixed-persona} if there exists a persona map
$\pi_p$, invariant across contexts, and a generation function $f_{\mathrm{LLM}}$, such that
\[
  \hat{Y}_{p,c}^{\mathcal{A}} = f_{\mathrm{LLM}}(\pi_p, X_c) + \eta_{p,c},
\]
where $X_c$ is the context input and $\eta_{p,c}$ is sampling noise independent of $\pi_p$ and
$X_c$.
\end{definition}

\begin{definition}[Person-context interaction residual]\label{def:interaction-residual}
For any variable $Z_{p,c}$,
\[
  R_{p,c}(Z) := Z_{p,c} - \bar{Z}_{p,\cdot} - \bar{Z}_{\cdot,c} + \bar{Z}_{\cdot,\cdot},
\]
the residual after removing person and context main effects, where $\bar Z_{p,\cdot}$ is the
average over contexts for fixed $p$, $\bar Z_{\cdot,c}$ is the average over persons for fixed $c$,
and $\bar Z_{\cdot,\cdot}$ is the grand average. We write
$R^{\mathcal{A}}_{p,c} := R_{p,c}(\hat Y^{\mathcal{A}})$ and
$R^{\mathrm{human}}_{p,c} := R_{p,c}(Y^{\mathrm{human}})$.
\end{definition}

\begin{definition}[Identity state decomposition and person-specific deviation]\label{def:regime-decomposition}
Let $\lambda_p$ denote person $p$'s soft identity-state membership (a mixture weight vector over
$k$ identity state components; a discrete label $S_p$ is the one-hot special case). Decompose the human
interaction residual as
\[
  R^{\mathrm{human}}_{p,c} = m(\lambda_p, c) + \Delta_{p,c}, \qquad
  m(\lambda_p,c) := \mathbb{E}[R^{\mathrm{human}}_{p,c} \mid \lambda_p, D_p, c],
\]
where $m(\lambda_p,c)$ is the \emph{state-typical shift} (predictable from identity state membership and
context) and $\Delta_{p,c}$ is the \emph{person-specific deviation} specific to person $p$. The notation suppresses the demographics $D_p$, which the shift may also depend on. Chameleon has no demographics, so $D_p$ is empty there. By
construction, $\mathbb{E}[\Delta_{p,c} \mid \lambda_p, D_p, c] = 0$.
\end{definition}

\subsection{Proposition 1: Representational Boundedness}

\begin{propos}\label{prop:boundedness}
Let $\mathcal{A}$ be any fixed-persona system (Definition~\ref{def:fixed-persona}). Substituting
$Z = \hat Y^{\mathcal{A}}$ into $R_{p,c}(Z)$ (Definition~\ref{def:interaction-residual}) and isolating
its deterministic part as $g(\pi_p,X_c)$ (the proof below derives $g$ explicitly by subtracting the marginal means of $f_{\mathrm{LLM}}$) gives
\[
  R^{\mathcal{A}}_{p,c} = g(\pi_p, X_c) + \eta_{p,c}
\]
for some function $g$, with $\mathrm{Cov}(g(\pi_p,X_c), \eta_{p,c}) = 0$. That is,
$R^{\mathcal{A}}_{p,c}$ depends on nothing outside $(\pi_p, X_c, \eta_{p,c})$: the design's
interaction structure is fully mediated by the persona representation and the context input.
\end{propos}

\begin{proof}
\textbf{Step 1 (set up the marginal means).} Define
\[
  \mu(\pi_p) := \mathbb{E}_{X_c}[f_{\mathrm{LLM}}(\pi_p, X_c)], \qquad
  \nu(X_c) := \mathbb{E}_{\pi_p}[f_{\mathrm{LLM}}(\pi_p,X_c)], \qquad
  \bar\mu := \mathbb{E}_{\pi_p,X_c}[f_{\mathrm{LLM}}(\pi_p,X_c)].
\]
$\mu(\pi_p)$ is persona $\pi_p$'s average simulated behavior across contexts (its ``person main
effect''); $\nu(X_c)$ is context $X_c$'s average simulated behavior across personas (its ``context
main effect''); $\bar\mu$ is the grand average.

\textbf{Step 2 (substitute the design's generative form into the residual).} By
Definition~\ref{def:fixed-persona}, $\hat Y^{\mathcal{A}}_{p,c} = f_{\mathrm{LLM}}(\pi_p,X_c) + \eta_{p,c}$. Substituting
into Definition~\ref{def:interaction-residual}'s residual formula:
\[
  R^{\mathcal{A}}_{p,c} = f_{\mathrm{LLM}}(\pi_p, X_c) + \eta_{p,c} - \mu(\pi_p) - \nu(X_c) + \bar{\mu}.
\]
(The marginal means of $\hat Y^{\mathcal{A}}$ reduce to $\mu(\pi_p)$, $\nu(X_c)$, $\bar\mu$ because
$\eta_{p,c}$ has mean zero and is independent of $\pi_p, X_c$, so it does not shift the marginals.)

\textbf{Step 3 (isolate the deterministic part).} Define
\[
  g(\pi_p, X_c) := f_{\mathrm{LLM}}(\pi_p, X_c) - \mu(\pi_p) - \nu(X_c) + \bar{\mu},
\]
a function of $(\pi_p,X_c)$ alone, containing no $\eta$ term. Then, from Step 2,
\[
  R^{\mathcal{A}}_{p,c} = g(\pi_p, X_c) + \eta_{p,c}.
\]

\textbf{Step 4 (independence of the two parts).} Since $\eta_{p,c}$ is independent of $\pi_p$ and
$X_c$ by Definition~\ref{def:fixed-persona}, and $g$ is a function only of $(\pi_p, X_c)$, it follows that
$\mathrm{Cov}(g(\pi_p,X_c),\eta_{p,c}) = 0$. Hence $R^{\mathcal{A}}_{p,c}$ depends on nothing beyond
$\pi_p$, $X_c$, and $\eta_{p,c}$.
\end{proof}

\noindent This mediation structure is what makes the design a \emph{channel}, since any dependence
between $R^{\mathcal{A}}_{p,c}$ and an external variable must route through $\pi_p$, $X_c$, or
$\eta_{p,c}$. Proposition~\ref{prop:non-transmission} uses this directly.

\subsection{Proposition 2: Person-Specific Non-Transmission}

\paragraph{Key Assumption.} $\pi_p \perp \Delta_{p,\cdot} \mid \lambda_p, D_p$ (i.e., $\pi_p$ and $\Delta_{p,\cdot}$ are independent given $\lambda_p, D_p$), where $D_p$ denotes
demographics, so persona construction is conditionally independent of person-specific deviation, given
identity state membership and demographics. Substantively, whatever process builds $\pi_p$, a bio, an
interview transcript, a categorical trait vector, conditions on who the person generally is, not
on how their behavior in person-specific ways shifts across contexts beyond what their identity state predicts.
This is defensible because $\Delta_{p,c}$ is a repeated-measures quantity, observable only from
behavior across multiple contexts, and standard persona-construction methods do not perform
multi-context behavioral elicitation.

\paragraph{Context Exogeneity Assumption.} $X_c \perp \Delta_{p,\cdot} \mid \lambda_p, D_p$ (i.e., $X_c$ and $\Delta_{p,\cdot}$ are independent given $\lambda_p, D_p$). Which context a person is observed or assigned in is conditionally independent of their person-specific
deviation, given identity state membership and demographics. Substantively, context assignment does not
select on a person's person-specific residual. Knowing a person's identity state and demographics, the context they end up in carries no further information about how they specifically deviate from
their identity state's typical response. This assumption is not equally plausible across designs, since it is close to exact for \emph{experimentally assigned} contexts, where context is randomized
independent of any respondent trait (\S\ref{sec:jmir}'s between-subject theme assignment), and
weaker for \emph{observationally sampled} contexts, where a person's person-specific tendencies could
influence which contexts they are observed in at all (\S\ref{sec:chameleon}'s subreddit choice is
plausibly self-selected in part on the very traits $\Delta_{p,c}$ captures). As with the Key
Assumption, exact independence is the strong case; the corollary below extends to its approximate
relaxation for both assumptions jointly.

\begin{propos}\label{prop:non-transmission}
Under Definition~\ref{def:fixed-persona}, Proposition~\ref{prop:boundedness}, independent sampling of persons, the Key Assumption, the Context Exogeneity Assumption, and the independence of persona and context given $(\lambda_p,D_p,\Delta_{p,\cdot})$,
\[
  \mathrm{Cov}\left(R^{\mathcal{A}}_{p,c}, \, \Delta_{p,c}\right) = 0, \quad \text{hence} \quad
  \mathrm{Cov}\left(R^{\mathcal{A}}_{p,c}, \, R^{\mathrm{human}}_{p,c}\right)
  = \mathrm{Cov}\left(R^{\mathcal{A}}_{p,c}, \, m(\lambda_p,c)\right).
\]
Any correlation a fixed-persona system achieves with real human interaction structure can
only come from matching the state-typical shift pattern, never from truly person-specific
identity activation.
\end{propos}

\begin{proof}
\textbf{Step 1 (conditional independence of the simulated residual).} Write $W := (\lambda_p, D_p)$. By
Proposition~\ref{prop:boundedness}, $R^{\mathcal{A}}_{p,c}$ is a function of $\pi_p$, $X_c$, $\eta_{p,c}$, and, through the
finite-sample marginals, of other persons' counterparts, which are independent of person
$p$'s own $\Delta_{p,c}$ because persons are sampled independently. The noise $\eta_{p,c}$ is independent of $\Delta_{p,c}$ (Setting).
The Key Assumption gives $\pi_p \perp \Delta_{p,c} \mid W$, and the Context Exogeneity Assumption gives
$X_c \perp \Delta_{p,c} \mid W$. Two pairwise independences do not give joint independence, so we also use the condition
from the Setting, $\pi_p \perp X_c \mid W, \Delta_{p,c}$. Then
$p(\pi_p, X_c \mid W, \Delta_{p,c}) = p(\pi_p \mid W, \Delta_{p,c})\,p(X_c \mid W, \Delta_{p,c}) = p(\pi_p \mid W)\,p(X_c \mid W)$,
so $(\pi_p, X_c) \perp \Delta_{p,c} \mid W$. Together with the independence of the noise and of other persons, this gives
$R^{\mathcal{A}}_{p,c} \perp \Delta_{p,c} \mid W$.

\textbf{Step 2 (apply the law of total covariance).}
\[
  \mathrm{Cov}(R^{\mathcal{A}}_{p,c}, \Delta_{p,c})
  = \mathbb{E}\left[\mathrm{Cov}(R^{\mathcal{A}}_{p,c}, \Delta_{p,c} \mid \lambda_p, D_p)\right]
  + \mathrm{Cov}\left(\mathbb{E}[R^{\mathcal{A}}_{p,c}\mid \lambda_p,D_p],\
    \mathbb{E}[\Delta_{p,c}\mid \lambda_p,D_p]\right).
\]

\textbf{Step 3 (both terms vanish).} The first term is zero by the conditional independence
established in Step 1. The second term is zero because $\mathbb{E}[\Delta_{p,c} \mid \lambda_p, D_p]
= 0$, which follows from $\mathbb{E}[\Delta_{p,c} \mid \lambda_p, D_p, c] = 0$ (Definition~\ref{def:regime-decomposition}) by averaging over $c$. So $\mathrm{Cov}(R^{\mathcal{A}}_{p,c},\Delta_{p,c}) = 0$.

\textbf{Step 4 (translate to the human residual).} Since
$R^{\mathrm{human}}_{p,c} = m(\lambda_p,c) + \Delta_{p,c}$,
\[
  \mathrm{Cov}(R^{\mathcal{A}}_{p,c}, R^{\mathrm{human}}_{p,c})
  = \mathrm{Cov}(R^{\mathcal{A}}_{p,c}, m(\lambda_p,c))
  + \mathrm{Cov}(R^{\mathcal{A}}_{p,c}, \Delta_{p,c})
  = \mathrm{Cov}(R^{\mathcal{A}}_{p,c}, m(\lambda_p,c)). \qedhere
\]
\end{proof}

\subsection{Robustness corollary}

We first record the single-assumption case, then extend it to cover both the Key Assumption and
the Context Exogeneity Assumption jointly, since the plausibility discussion above notes the latter
is the weaker of the two for observationally sampled contexts.

\begin{corol}[Single-assumption relaxation]\label{cor:robustness-single}
We formalize approximate independence directly, rather than treating it as following from some
looser notion of ``approximately holds'': $\pi_p$ and $\Delta_{p,\cdot}$ are
\emph{$\epsilon$-independent} given $(\lambda_p,D_p)$ if
$I(\pi_p; \Delta_{p,\cdot} \mid \lambda_p, D_p) \leq \epsilon$; this recovers the Key Assumption
exactly when $\epsilon=0$. Suppose the Context Exogeneity Assumption and the independence of persona and context given $(\lambda_p,D_p,\Delta_{p,\cdot})$ hold exactly. Then $I((\pi_p,X_c);\Delta_{p,c}\mid\lambda_p,D_p)=I(\pi_p;\Delta_{p,c}\mid\lambda_p,D_p)\le\epsilon$. By Proposition~\ref{prop:boundedness}, $R^{\mathcal{A}}_{p,c}$ is a noisy function of $(\pi_p,X_c)$, so the data processing inequality gives $I(R^{\mathcal{A}}_{p,c};\Delta_{p,c}\mid\lambda_p,D_p)\le\epsilon$. By Pinsker's inequality the
total variation distance between the joint and product-of-marginals distributions of
$(R^{\mathcal{A}}_{p,c},\Delta_{p,c})$ given $(\lambda_p,D_p)$ is bounded by $\sqrt{\epsilon/2}$. With
$|R^{\mathcal{A}}_{p,c}|, |\Delta_{p,c}| \leq M$ almost surely (a reasonable assumption for bounded
Likert-scale outcomes), covariance is bounded by total variation distance times the product of
ranges, giving
\[
  \left|\mathrm{Cov}(R^{\mathcal{A}}_{p,c}, \Delta_{p,c})\right| \leq 4M^2\sqrt{\epsilon/2}.
\]
This degrades gracefully to a small-but-nonzero bound rather than an exact zero when the Key
Assumption is only approximately true, the more realistic case for richer persona representations.
\end{corol}

\begin{corol}[Joint relaxation]\label{cor:robustness-joint}
We extend Corollary~\ref{cor:robustness-single}'s definition to both assumptions. Suppose
$\pi_p$ is $\epsilon_1$-independent of $\Delta_{p,\cdot}$ given $(\lambda_p,D_p)$, that is $I(\pi_p; \Delta_{p,\cdot} \mid \lambda_p, D_p) \leq \epsilon_1$, recovering the Key Assumption at $\epsilon_1=0$, and $X_c$ is $\epsilon_2$-independent of $\Delta_{p,\cdot}$ given $(\lambda_p,D_p)$, that is $I(X_c; \Delta_{p,\cdot} \mid \lambda_p, D_p) \leq \epsilon_2$, recovering the Context Exogeneity
Assumption at $\epsilon_2=0$. By the chain rule for conditional mutual information,
\[
  I\big((\pi_p,X_c); \Delta_{p,c} \mid \lambda_p, D_p\big)
  \;=\; I(\pi_p;\Delta_{p,c}\mid\lambda_p,D_p)
  \;+\; I(X_c;\Delta_{p,c}\mid\pi_p,\lambda_p,D_p).
\]
The first term on the right is at most $\epsilon_1$ by assumption. The second term is not directly
bounded by $\epsilon_2$, which bounds $I(X_c;\Delta_{p,c}\mid\lambda_p,D_p)$ without conditioning on $\pi_p$, so we invoke one further simplification, stated explicitly rather than
folded into the derivation.

\paragraph{Cross-Independence Simplification.} $I(X_c;\Delta_{p,c}\mid\pi_p,\lambda_p,D_p) \approx
I(X_c;\Delta_{p,c}\mid\lambda_p,D_p) \leq \epsilon_2$. This holds exactly when $\pi_p$ is
conditionally independent of the pair $(X_c,\Delta_{p,c})$ given $(\lambda_p,D_p)$. This is plausible because $\pi_p$ is, by Definition~\ref{def:fixed-persona}, invariant across contexts and therefore
carries no context-indexed information beyond what $\lambda_p,D_p$ already summarize; conditioning
on it should not materially change how much $X_c$ reveals about $\Delta_{p,c}$. This is an
additional, separately falsifiable claim, not a consequence of the two named assumptions above, and
we flag it as such, since it is more plausible when persona construction and context assignment are
themselves independent processes (true by design in \S\ref{sec:jmir}'s experimental assignment;
requires separate justification wherever context is observationally sampled rather than assigned).

Under this simplification,
\[
  I\big((\pi_p,X_c); \Delta_{p,c} \mid \lambda_p, D_p\big) \leq \epsilon_1 + \epsilon_2 =: \epsilon.
\]
By Proposition~\ref{prop:boundedness}, $R^{\mathcal{A}}_{p,c} = g(\pi_p,X_c) + \eta_{p,c}$ with $g$ a
deterministic function of $(\pi_p,X_c)$ and $\eta_{p,c}$ independent of $\Delta_{p,c}$; the data
processing inequality then gives
$I(R^{\mathcal{A}}_{p,c};\Delta_{p,c}\mid\lambda_p,D_p) \leq I\big((\pi_p,X_c);\Delta_{p,c}\mid
\lambda_p,D_p\big) \leq \epsilon$. Applying Pinsker's inequality and the same bounded-outcome
argument as Corollary~\ref{cor:robustness-single},
\[
  \left|\mathrm{Cov}(R^{\mathcal{A}}_{p,c}, \Delta_{p,c})\right| \leq 4M^2\sqrt{\epsilon/2}
  = 4M^2\sqrt{(\epsilon_1+\epsilon_2)/2}.
\]
As with the single-assumption case, this degrades gracefully rather than collapsing, since the bound scales with the combined violation of both assumptions, and remains small whenever both
$\epsilon_1$ and $\epsilon_2$ are small, which is the expected case for closed-schema personas under
experimentally assigned contexts, and the case requiring the most caution for rich, open-ended
personas under observationally sampled ones.
\end{corol}

\subsection{Assumption defensibility across the fixed-persona family}

Each assumption is more plausible for some designs than for others. The Key Assumption is strongest for closed-schema personas built from fixed demographic or categorical fields, which have no channel for person-specific cross-context deviation. It is weaker for rich, open-ended personas such as interview transcripts, which could incidentally encode traits correlated with person-specific activation. There Corollary~\ref{cor:robustness-single} or Corollary~\ref{cor:robustness-joint} carries the argument in place of the exact-zero result, and the degradation is graceful (small $\epsilon_1$, small bound). The Context Exogeneity Assumption is strongest for experimentally assigned contexts, such as the between-subject theme randomization in \S\ref{sec:jmir}, and weaker for observationally sampled ones, such as the subreddit choices in \S\ref{sec:chameleon}, where a person's tendencies could influence which contexts they appear in. Corollary~\ref{cor:robustness-joint} carries the argument there, and \S\ref{sec:criterion-scope} states this as a scope condition.

\section{Identity-State Specifications and Empirical Methodology}
\label{sec:regime-specs}

This appendix collects the construction detail, calibration history, and supporting validation
for the two empirical demonstrations in \S\ref{sec:chameleon} and \S\ref{sec:jmir}. None of this
detail is needed to evaluate the paper's central claims, which are stated in full in the main
text. It is provided for reproducibility and for readers who want to check the specific
engineering decisions behind Arm C in each domain. We call a specification \emph{locked} when it was fixed in writing before the run it governs.

Table~\ref{tab:cross-domain-synthesis} summarizes both domains' main results side by side,
organized by arm rather than by domain, with the reading each result supports.

\begin{table}[t]
  \centering
  \small
  \caption{Both domains' main results, by arm, with interpretation. Person-specific
  signal is near zero for every arm in Chameleon, and the test cannot separate the arms at that
  size (Appendix~\ref{sec:chameleon-full-test}). The motivational-themes study has one observation
  per person, so it tests the aggregate interaction only (\S\ref{sec:empirical-interpretation}).}
  \label{tab:cross-domain-synthesis}
  \begin{tabular}{p{0.24\linewidth}p{0.14\linewidth}p{0.14\linewidth}p{0.14\linewidth}p{0.22\linewidth}}
    \toprule
    Metric & Arm A & Arm B & Arm C & Reading \\
    \midrule
    Chameleon ICC (real $\approx$0.013) & 0.909 & 0.799 & 0.285 & Arm C closer in aggregate but uneven by construct (Table~\ref{tab:chameleon-icc}) \\
    Chameleon person-specific correlation, significant constructs (of 5) & 3 & 2 & 1 & Near zero for every arm, and the test cannot separate them below about 0.04 (Table~\ref{tab:chameleon-idio}) \\
    Motivational-themes trust main effect & n/a & n/a & reproduced precisely & Generative step checked directly (\S\ref{sec:jmir-results}) \\
    Motivational-themes 6-coefficient direction match & 5/6 & 0/6 & 2/6 domain-general, 1/6 theme-amplified & Arm C's gap lies in how $\pi_I$ selects the state and the generative step works, so it can be closed in principle, unlike Arm A's ceiling (Table~\ref{tab:jmir-specs}) \\
    \bottomrule
  \end{tabular}
\end{table}

\subsection{Full Chameleon Empirical Test}
\label{sec:chameleon-full-test}

The target population is 1,667 Reddit users (Chameleon \cite{harry2026acl}), each with three posts across distinct subreddits, scored on five Big Five personality constructs plus twenty-one auxiliary trait dimensions using SEANCE, a psycholinguistic text-analysis tool that scores language for psychological content. The real signal is faint. Bias-corrected ICC(1) is approximately 0.01 to 0.02 across all five constructs, so 98 to 101\% of real variance is within-person, with the upper end above 100\% because Neuroticism's bias-corrected estimate is slightly negative ($-0.007$, Table~\ref{tab:chameleon-icc}), a known property of this estimator when the true value is near zero. Real person-context interaction variance is small in absolute terms, at 0.072 for extraversion, 0.049 for agreeableness, 0.067 for conscientiousness, 0.022 for neuroticism, and 0.043 for openness.

Each user's own three real (user, subreddit) cells are simulated under the three arms of \S\ref{sec:shared-design}. Arm A's persona is a bio built from the user's twenty-one non-Big-Five dimensions (Schwartz Values, Self-Determination Theory, and the DOSPERT risk-taking scale), and the Big Five is excluded because it is the outcome, so that the model cannot echo the answer. Arm B adds the subreddit name. Arm C adds the active identity state (Appendix~\ref{sec:regime-spec-chameleon}). The full-scale run covers all 1,667 users with two or three replicates per cell, in the hybrid reasoning-and-rating format of Appendix~\ref{sec:elicitation-format-validation}.

\paragraph{Results.}
\label{sec:chameleon-results}

\emph{ICC.} Real ICC clusters in a narrow, near-zero band (0.01 to 0.02). Arms A and B invert this near-completely, with between-person shares of 83 to 95\% for Arm A (mean 0.909) and 60 to 89\% for Arm B (mean 0.799) against a real mean of 0.013, as \S\ref{sec:incapability} predicts. Arm C's aggregate ICC (mean 0.285) sits substantially closer to the real value, but the improvement is not uniform (Table~\ref{tab:chameleon-icc}). Extraversion is recovered almost exactly, while agreeableness, conscientiousness, and openness overshoot by roughly 24 to 65-fold, and neuroticism overshoots more strongly in the negative direction. A person-in-context design that had recovered the real structure would show similarly small ICC on every construct, whereas Arm C's values spread widely across constructs.

\begin{table}[t]
  \centering
  \caption{Between-person variance share (bias-corrected ICC) by construct, real population vs.\ each arm. Real ICC clusters near zero across all five constructs. Arms A and B invert this near-completely. Bold marks the arm closest to the real value, which is Arm C in every row, though it is far above the real value on four of the five constructs.}
  \label{tab:chameleon-icc}
  \begin{tabular}{lrrrr}
    \toprule
    Construct & Real & Arm A & Arm B & Arm C \\
    \midrule
    Extraversion      & 0.019  & 0.954 & 0.861 & \textbf{0.023} \\
    Agreeableness     & 0.009  & 0.933 & 0.891 & \textbf{0.574} \\
    Conscientiousness & 0.022  & 0.901 & 0.800 & \textbf{0.585} \\
    Neuroticism       & $-$0.007 & 0.828 & 0.596 & \textbf{$-$0.248} \\
    Openness          & 0.020  & 0.930 & 0.846 & \textbf{0.489} \\
    \midrule
    Mean              & 0.013  & 0.909 & 0.799 & \textbf{0.285} \\
    \bottomrule
  \end{tabular}
\end{table}

\emph{Person-specific signal.} This is the direct test of Proposition 2, computed across approximately 4,568 person-context comparisons per construct (Table~\ref{tab:chameleon-idio}). The mean correlation is 0.026 for Arm A, 0.025 for Arm B, and 0.004 for Arm C, near zero for every arm. Arm A is significant on three of five constructs (extraversion, conscientiousness, neuroticism), Arm B on two, and Arm C on one (extraversion). Small positive values for the fixed-persona arms are what the robustness corollary expects. Arm C's point estimates are negative on agreeableness and conscientiousness, but its intervals include zero, as do all arms' intervals on openness. The detectability analysis below shows that the test cannot separate the arms at this size.

\begin{table}[t]
  \centering
  \small
  \caption{Person-specific residual correlation (state-adjusted shift correlation with $\Delta_{p,c}$) by trait and arm, with 95\% bootstrap intervals in brackets (users resampled). An asterisk marks an interval that excludes zero. The mean is over the five traits.}
  \label{tab:chameleon-idio}
  \begin{tabular}{lccc}
    \toprule
    Trait & Arm A & Arm B & Arm C \\
    \midrule
    Extraversion      & 0.039$^{*}$ [0.012, 0.065] & 0.064$^{*}$ [0.035, 0.092] & 0.030$^{*}$ [0.004, 0.057] \\
    Agreeableness     & 0.010 [$-$0.020, 0.038] & 0.012 [$-$0.018, 0.042] & $-$0.020 [$-$0.050, 0.013] \\
    Conscientiousness & 0.033$^{*}$ [0.005, 0.060] & 0.012 [$-$0.016, 0.041] & $-$0.005 [$-$0.032, 0.021] \\
    Neuroticism       & 0.035$^{*}$ [0.003, 0.065] & 0.028$^{*}$ [0.002, 0.055] & 0.004 [$-$0.020, 0.027] \\
    Openness          & 0.011 [$-$0.019, 0.042] & 0.008 [$-$0.023, 0.038] & 0.011 [$-$0.019, 0.041] \\
    \midrule
    Mean              & 0.026 & 0.025 & 0.004 \\
    \bottomrule
  \end{tabular}
\end{table}

\emph{Detectability.} Each person-specific correlation rests on about 4,570 person-context comparisons per construct, and its interval comes from a bootstrap that resamples whole users (1,000 resamples). The standard errors implied by those intervals are 0.012 to 0.016 (mean 0.0145), so an observed correlation must be about 0.033 to 0.045 (mean 0.041) to be detected with 80\% power. A simulation that gives a synthetic system a known correlation with the real person-specific residuals confirms this, with detection rates of 78\% at 0.04 and 98\% at 0.05 (60 repetitions each). Arm A's and Arm B's mean correlations (0.026 and 0.025) lie below that size, and Arm C's mean (0.004) lies far below it. The test therefore rules out an observed correlation well above 0.04 in every arm and cannot separate the arms below about 0.04. The real residual also contains post-level noise, so even a perfect simulator would score well below 1 on this measure, and a near-zero value does not show that person-specific signal is absent.

\subsection{Identity-State Specification: Chameleon}
\label{sec:regime-spec-chameleon}

Arm C's set of identity states $\mathcal{I}_p$ was built data-first. A six-component Gaussian mixture model
was fit to the 1,667 users' twenty-six-dimensional trait vectors (five Big Five constructs plus
twenty-one auxiliary dimensions). Psychological interpretation of each component was applied only
after fitting, through three independent blind literature-grounding panels that did not have
access to each other's labels until convergence was checked.

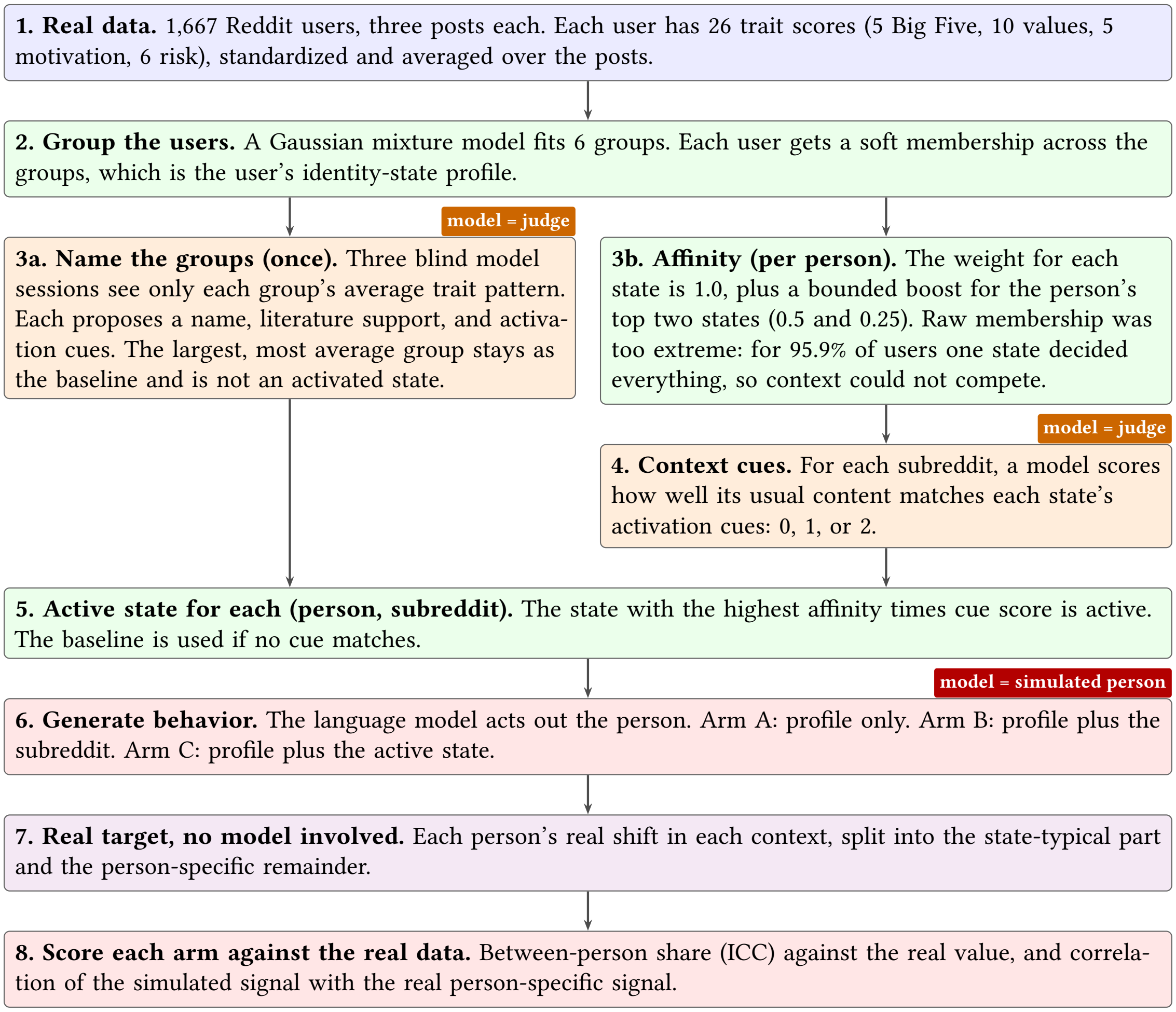
\begin{figure}[t]
  \centering
  \input{figures/fig_appendix_chameleon_pipeline.tex}
  \caption{The Chameleon pipeline, from the raw trait scores to how each arm is scored against real people. The mixture model's soft membership feeds two separate jobs: one-time naming of the identity states (Appendix~\ref{sec:regime-spec-chameleon}) and per-person affinity, which combines with context cues to pick the active state for each (person, subreddit) cell. Orange tags mark where the language model acts as judge, and the red tag marks where it acts as the simulated person.}
  \label{fig:chameleon-pipeline}
\end{figure}

The panels converged on treating the largest, least-differentiated component (40.3\% weight) as
an unmarked statistical baseline rather than a distinct activated identity state, in the spirit of
Fleeson's view of a trait as the center of a distribution of states (\S\ref{sec:taxonomy}). The state-typical shift term $m(\lambda_p,c)$
excludes this baseline component. The remaining five components, ranging from 2.7\% to 21.4\%
weight, carry Arm C's context-conditional behavior. Table~\ref{tab:chameleon-regimes} names and
interprets all six. Figure~\ref{fig:chameleon-pipeline} diagrams the full pipeline below, from the
raw trait vectors through to how each arm is scored against real people.

\begin{table}[t]
  \centering
  \caption{All six GMM components, named and interpreted by the three blind literature-grounding
  panels. Component 2 is the unmarked baseline (excluded from $m(\lambda_p,c)$), and the remaining five
  carry Arm C's context-conditional behavior. Confidence reflects panel agreement on whether the
  component reflects a real trait profile or a possible confound.}
  \label{tab:chameleon-regimes}
  \small
  \begin{tabular}{p{0.19\linewidth}p{0.1\linewidth}p{0.6\linewidth}}
    \toprule
    Component & Weight & Interpretation \\
    \midrule
    2: Conventional Growth-Oriented (baseline) & 40.3\% & Mild achievement/self-direction elevation, domain-specific risk aversion. Not an activated identity state. \\
    0: Expansive Engaged & 21.4\% & Broad elevation across nearly all variables, flagged by all three panels as a possible general verbosity/lexical-elevation confound and not fully distinguishable from a real trait profile with this data. \\
    4: Quiet Withdrawn & 18.1\% & Broadly suppressed but neuroticism flat: low-approach-motivation withdrawal, not anxious withdrawal. \\
    5: Conventional Risk-Tolerant & 10.9\% & Elevated ethical/health-safety/gambling risk-taking alongside retained tradition-value commitment. \\
    1: Amotivated Disinhibited & 6.6\% & Most extreme component, low confidence: flagged by all three panels as possible outlier absorption in the mixture rather than a stable subpopulation. \\
    3: High-Intensity Reactive & 2.7\% & Smallest component, treated as diagnostic rather than pre-judged, due to an atypical co-elevation pattern also possibly reflecting an extraction confound (bootstrap-stable across resamples, so not a sampling artifact, but a measurement-validity question remains open). \\
    \bottomrule
  \end{tabular}
\end{table}

Table~\ref{tab:chameleon-wording} gives the wording that Arm C passes to the model for each activated state. The baseline component has no wording, since it is not an activated state.

\begin{table}[t]
  \centering
  \small
  \caption{Wording of the five activated Chameleon identity states, as given to the model. The active state appears as ``Primarily [name] ([wording])'', and a second state, if any, as ``with some influence from being [name] ([wording])''.}
  \label{tab:chameleon-wording}
  \begin{tabular}{p{0.22\linewidth}p{0.70\linewidth}}
    \toprule
    State & Wording \\
    \midrule
    Expansive Engaged & socially confident and broadly engaged across many areas of life, comfortable taking risks in multiple domains \\
    Amotivated Disinhibited & disengaged and low-investment in most goals, but willing to bend rules around safety, ethics, or gambling \\
    High-Intensity Reactive & emotionally intense and highly driven, engaged and a bit anxious at the same time \\
    Quiet Withdrawn & low-key and withdrawn, not very focused on achievement or social striving, calm rather than anxious \\
    Conventional Risk-Tolerant & values tradition and convention, but willing to take risks around ethics, safety, or gambling despite that \\
    \bottomrule
  \end{tabular}
\end{table}

\paragraph{Calibration history.} The initial construction used each user's raw GMM
posterior $\lambda_p$ directly as the activation signal, which failed because posteriors are extreme by
construction (95.9\% of users have over 90\% of their non-baseline mass on a single identity state),
leaving context almost no room to compete with a person's default identity state. Context-conditioned and
static top-2 identity state selections were identical in 97.3\% of piloted cells.

The fix separates \emph{affinity} (which identity states are available to a person, drawn from
$\lambda_p$) from \emph{activation} (which identity state is realized in a specific context). This adopts
the same weight-times-cue-match structure independently arrived at for the motivational-themes
study's identity state competition (below), in which a modest affinity weight $w_{p,r}$ is combined with a
context-only cue-match score $a_{r,c}$, and the realized activation is
$\arg\max_r\, w_{p,r} \cdot a_{r,c}$.

\subsection{Identity-State Specification: Motivational-Themes Study}
\label{sec:regime-spec-jmir}

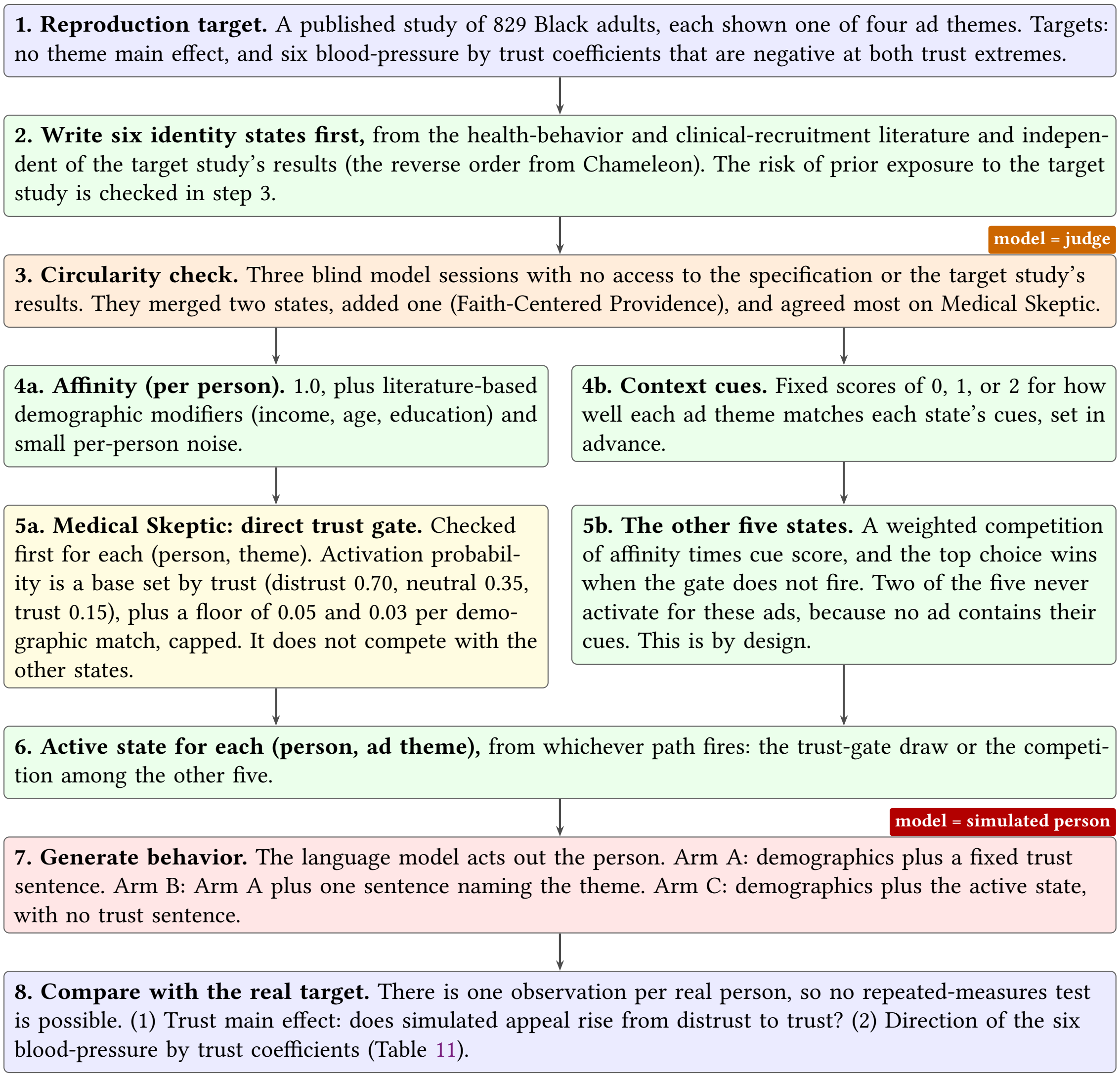
\begin{figure}[t]
  \centering
  \input{figures/fig_appendix_jmir_pipeline.tex}
  \caption{The motivational-themes pipeline. Medical Skeptic skips the shared competition through a direct trust gate that is checked first for each (person, theme) cell. The other five states compete only if that gate does not fire. Orange tags mark where the language model acts as judge, and the red tag marks where it acts as the simulated person.}
  \label{fig:jmir-pipeline}
\end{figure}

Arm C's set of identity states $\mathcal{I}_p$ here was built literature-first, the inverse of Chameleon's
data-first approach. Six theory-grounded identity states were specified from the health-behavior and
clinical-trial-recruitment literature, independent of the target study's own findings. Two of the
six identity states never activate for this study's four stimuli specifically, by intended construction,
since not every identity state in the set is relevant to every domain the set might be reused
in. Figure~\ref{fig:jmir-pipeline} diagrams the full pipeline, including Medical Skeptic's direct
trust gate and how the remaining five identity states compete for activation.

\paragraph{Circularity check.} Because this specification could not rule out prior exposure to the target study's findings, an independent circularity check was run before locking the taxonomy. Three isolated LLM sessions were each given only a neutral task description (build a literature-grounded identity state taxonomy for this population and context) with no access to this specification or the target study's findings, and instructed to disregard the target study if it surfaced in their own search. One session had zero exposure to the study across its entire search. Their independently built taxonomies were compared against the draft specification for convergence, driving two merges, one addition (an identity state that emerged unprompted in all three), and giving Medical Skeptic, the identity state the paper's reproduction target depends on, the strongest independent convergence of all six.

Medical Skeptic is the identity state the reproduction target theoretically depends on. It is
driven primarily by trust tertile, with base activation rates of 0.70 for distrust, 0.35 for
neutral, and 0.15 for trust, plus capped secondary demographic correlates that are constrained so
they cannot outvote trust itself.

Table~\ref{tab:jmir-states} gives the wording and the affinity rules for all six states. A person's affinity for a state is $1$ plus the sum of the bumps for the demographic categories they belong to. When the Medical Skeptic gate does not fire, the state with the highest affinity times cue score among the other five is active. Medical Skeptic's activation probability is the trust base rate plus a constant 0.05 for the institutional framing present in every ad, plus 0.03 for each of three secondary correlates (rural residence, age 60 or older, a bachelor's degree or higher), clipped to the range 0.05 to 0.95.

\begin{table}[t]
  \centering
  \small
  \caption{The six identity states of the motivational-themes study. The wording follows ``Right now, this person's salient concern is:'' in the prompt. Affinity bumps are added to a base affinity of 1.}
  \label{tab:jmir-states}
  \begin{tabular}{p{0.15\linewidth}p{0.52\linewidth}p{0.25\linewidth}}
    \toprule
    State & Wording & Affinity bumps \\
    \midrule
    Community \& Scientific Contribution & You care about contributing to knowledge or to your community. You weigh whether the science is credible and well-reviewed, or whether this specific effort will meaningfully benefit people like you. & +0.6 bachelor's degree or higher; +0.2 age 40 to 59 \\
    Health Focused & You are focused on your own health and whether this could directly improve your condition. You care about clinical relevance and access to results relevant to your own health. & +0.6 blood-pressure diagnosis; +0.2 age 50 or older \\
    Medical Skeptic & You are wary of medical institutions and clinical research. You scrutinize who is running this, whether Black investigators are involved, and whether this could be another instance of institutions using people like you without giving back. & Trust gate: 0.70 (distrust), 0.35 (neutral), 0.15 (trust) \\
    Prudential Consumer & You weigh this mainly as a practical exchange: is the compensation or incentive worth the time and effort involved, given your financial situation. & +0.6 income under \$35,000; +0.2 rural \\
    Privacy \& Surveillance Wary & You are concerned about who gets your data, how it will be shared, and whether digital or online aspects of this study put your privacy at risk. & +0.6 age 60 or older; +0.2 rural \\
    Faith-Centered Providence & You think about health decisions partly through your faith, including prayer, stewardship of your body, and whether this aligns with your spiritual life, alongside the practical details. & +0.6 age 60 or older; +0.2 rural \\
    \bottomrule
  \end{tabular}
\end{table}

Table~\ref{tab:jmir-cues} gives the cue scores for the four ads, which are fixed numbers set in the code. The Medical Skeptic entry applies to the earlier shared competition (versions 1.1 to 1.3). Under the trust gate that replaced it, institutional framing adds a constant 0.05 to its activation probability for every ad.

\begin{table}[t]
  \centering
  \small
  \caption{Cue scores (0 to 2) of each ad theme for each identity state, set in the code. Scores of 0 are omitted.}
  \label{tab:jmir-cues}
  \begin{tabular}{lp{0.60\linewidth}}
    \toprule
    Ad theme & Identity states with a nonzero cue score \\
    \midrule
    Science (research) & Community \& Scientific Contribution 2, Medical Skeptic 2 \\
    Community & Community \& Scientific Contribution 2, Medical Skeptic 2 \\
    Blood pressure & Health Focused 2, Medical Skeptic 2 \\
    Perks (compensation) & Prudential Consumer 2, Medical Skeptic 2 \\
    \bottomrule
  \end{tabular}
\end{table}

\paragraph{Calibration history.} Arm C's mechanism required four locked revisions before passing
internal-validity checks, each triggered by a specific diagnostic failure rather than
outcome-directed tuning. The first implementation, a literature-cited competition among all six
identity states, was not self-calibrating, since individually well-grounded modifier magnitudes let the intended
trait be outvoted by unrelated, more prevalent correlates (a college-education correlate common to
76\% of the pilot personas overwhelmed the trust-distrust modifier defining Medical Skeptic). The
mechanism that finally passed instead special-cases Medical Skeptic as a direct function of its
own defining trait.

\paragraph{The theme-amplified refinement.} Locked before evaluation, this refinement
amplifies the trust-linked identity state's activation spread specifically for the blood-pressure theme.
It is reported in the main text as a weaker test than the domain-general mechanism, since the
amplified theme was chosen with knowledge of where the real effect is located. In both domains, literature or statistical grounding alone did not guarantee that a mechanism let context and person jointly determine behavior, so each mechanism was checked directly with the diagnostics above.

\paragraph{Model-based steps.} In Chameleon, the context cue-matching score $a_{r,c}$ is produced by Claude Haiku 4.5, the same language model that generates persona behavior, in a separate call. This is an LLM-as-judge design in the sense of \citet{zheng2023judging} and \citet{gu2024survey}. In the motivational-themes study the cue scores are fixed integers from 0 to 2 for each ad theme (Table~\ref{tab:jmir-cues}), set in the code from the specification's activation cues. The Medical Skeptic score was raised from 1 to 2 after the first pilot showed that the state never activated, and the state was later replaced by a direct trust gate. The literature-grounding and circularity-check panels described above and in Appendices~\ref{sec:regime-spec-chameleon} and \ref{sec:regime-spec-jmir} were isolated agent sessions of Claude Sonnet 5.5 run through Claude Code, with no access to each other's output. They were not human raters. \S\ref{sec:discussion-limitations} discusses the resulting judge-and-subject concern.

\subsection{Motivational-Themes Study: Analysis Specification and Robustness}
\label{sec:jmir-robustness}

\paragraph{Published target.} Table~\ref{tab:jmir-target} lists the design and results of the study that serve as the target in \S\ref{sec:jmir}.

\begin{table}[t]
  \centering
  \caption{Design and published results of the motivational-themes study \cite{harry2026jmir} that serve as the target. Coefficients are the blood-pressure by trust interactions in log-odds, relative to neutral trust and to the contribution-to-science theme, with $p$-values.}
  \label{tab:jmir-target}
  \small
  \begin{tabular}{p{0.28\linewidth}p{0.62\linewidth}}
    \toprule
    Respondents & 829 Black adults in the United States \\
    Design & Between-subject. Each respondent saw one of four ad themes \\
    Themes & Contribution to science, community benefit, blood-pressure management, financial compensation \\
    Outcomes & Appeal, willingness to participate, willingness to recommend (5-point scales) \\
    Moderator & Trust in the healthcare system (distrust, neutral, trust) \\
    Model & Ordered logit with theme by trust interaction, demographic adjusters, and a random intercept for repeated measures \\
    Theme main effects & None significant on any outcome ($p=.25$ to $.93$) \\
    \midrule
    \multicolumn{2}{l}{Blood-pressure by trust interactions} \\
    Appeal, trust side & $-0.87$ ($p=.06$) \\
    Appeal, distrust side & $-0.92$ ($p=.04$) \\
    Participate, trust side & $-0.88$ ($p=.04$) \\
    Participate, distrust side & $-0.70$ ($p=.11$) \\
    Recommend, trust side & $-0.56$ ($p=.20$) \\
    Recommend, distrust side & $-1.03$ ($p=.02$) \\
    \bottomrule
  \end{tabular}
\end{table}

\paragraph{What was compared.} The real study reports six blood-pressure by trust coefficients, one for each of three outcomes (appeal, willingness to participate, willingness to recommend) at each trust extreme, relative to neutral trust and to the contribution-to-science theme. For each simulated arm we estimate the same six interactions and record whether the sign matches the published sign. All six published coefficients are negative, so direction-uncorrelated guessing matches 3 of 6 on average. Table~\ref{tab:jmir-coefs} lists the six simulated coefficients for each arm under the main-text specification.

\paragraph{Specification choice.} Before computing any arm's results, we fixed one primary specification to mirror the original study, an ordered logit (a cumulative-link model) for each outcome, with theme and trust interacted, the contribution-to-science theme and neutral trust as baselines, and the demographic adjusters of the published model. The original model has a random intercept for repeated measures. We replace it with standard errors clustered by persona. The main text reports a linear version of the same model, which gives contrasts in rating points, for the following reason. Simulated ratings follow the trust tier almost deterministically. In Arm A's appeal ratings the fitted trust effects reach $+6.4$ and $-11.9$ log-odds, which is close to separation, and the ordered-logit interaction coefficients are then unstable in size. A shift of $-0.12$ rating points appears as $-1.7$ log-odds. Sign counts are the same under both models for the baseline we use (Table~\ref{tab:jmir-specs}), so this choice does not change the direction results. With 6{,}632 personas almost every coefficient is statistically significant, so we compare direction and size and give significance only for reference.

\begin{table}[t]
  \centering
  \caption{Direction matches (of 6) under seven analysis specifications. Rows 1 to 4 use the contribution-to-science theme as the baseline, as the original study does. Rows 5 to 7 compare blood pressure with the pooled other three themes. Arm A leads under the original baseline, and the amplified refinement leads only under the pooled baseline.}
  \label{tab:jmir-specs}
  \small
  \begin{tabular}{p{0.46\linewidth}rrrr}
    \toprule
    Specification & A & B & Domain-general & Amplified \\
    \midrule
    1. Ordered logit, demographics (pre-specified) & 5 & 0 & 2 & 2 \\
    2. Linear, demographics (main text)            & 5 & 0 & 2 & 1 \\
    3. Ordered logit, no demographics              & 5 & 0 & 1 & 2 \\
    4. Linear, no demographics                     & 5 & 0 & 1 & 1 \\
    5. Ordered logit, pooled baseline, demographics & 2 & 0 & 1 & 3 \\
    6. Linear, pooled baseline, demographics        & 3 & 0 & 1 & 3 \\
    7. Ordered logit, pooled baseline, no demographics & 2 & 0 & 1 & 2 \\
    \bottomrule
  \end{tabular}
\end{table}

\begin{table}[t]
  \centering
  \caption{Specification 2 (main text): blood-pressure by trust interaction, in rating points, for each arm. Published values are log-odds with $p$-values, so sizes are not on a shared scale. Asterisks mark simulated coefficients with $p<.05$ (standard errors clustered by persona). Direction matches are the entries with a negative sign.}
  \label{tab:jmir-coefs}
  \small
  \begin{tabular}{llrrrrr}
    \toprule
    Outcome & Trust side & Published ($p$) & A & B & Domain-general & Amplified \\
    \midrule
    Appeal      & Trust    & $-0.87$ (.06) & $-0.13^{*}$ & $+0.16^{*}$ & $-0.01$ & $+0.14^{*}$ \\
    Appeal      & Distrust & $-0.92$ (.04) & $+0.01$ & $+0.42^{*}$ & $+0.28^{*}$ & $+0.04$ \\
    Participate & Trust    & $-0.88$ (.04) & $-0.34^{*}$ & $+0.29^{*}$ & $-0.01$ & $+0.14^{*}$ \\
    Participate & Distrust & $-0.70$ (.11) & $-0.04$ & $+0.33^{*}$ & $+0.24^{*}$ & $+0.09$ \\
    Recommend   & Trust    & $-0.56$ (.20) & $-0.20^{*}$ & $+0.30^{*}$ & $+0.09$ & $+0.41^{*}$ \\
    Recommend   & Distrust & $-1.03$ (.02) & $-0.08^{*}$ & $+0.33^{*}$ & $+0.18$ & $-0.18^{*}$ \\
    \bottomrule
  \end{tabular}
\end{table}

\begin{figure}[t]
  \centering
  \includegraphics[width=0.90\linewidth]{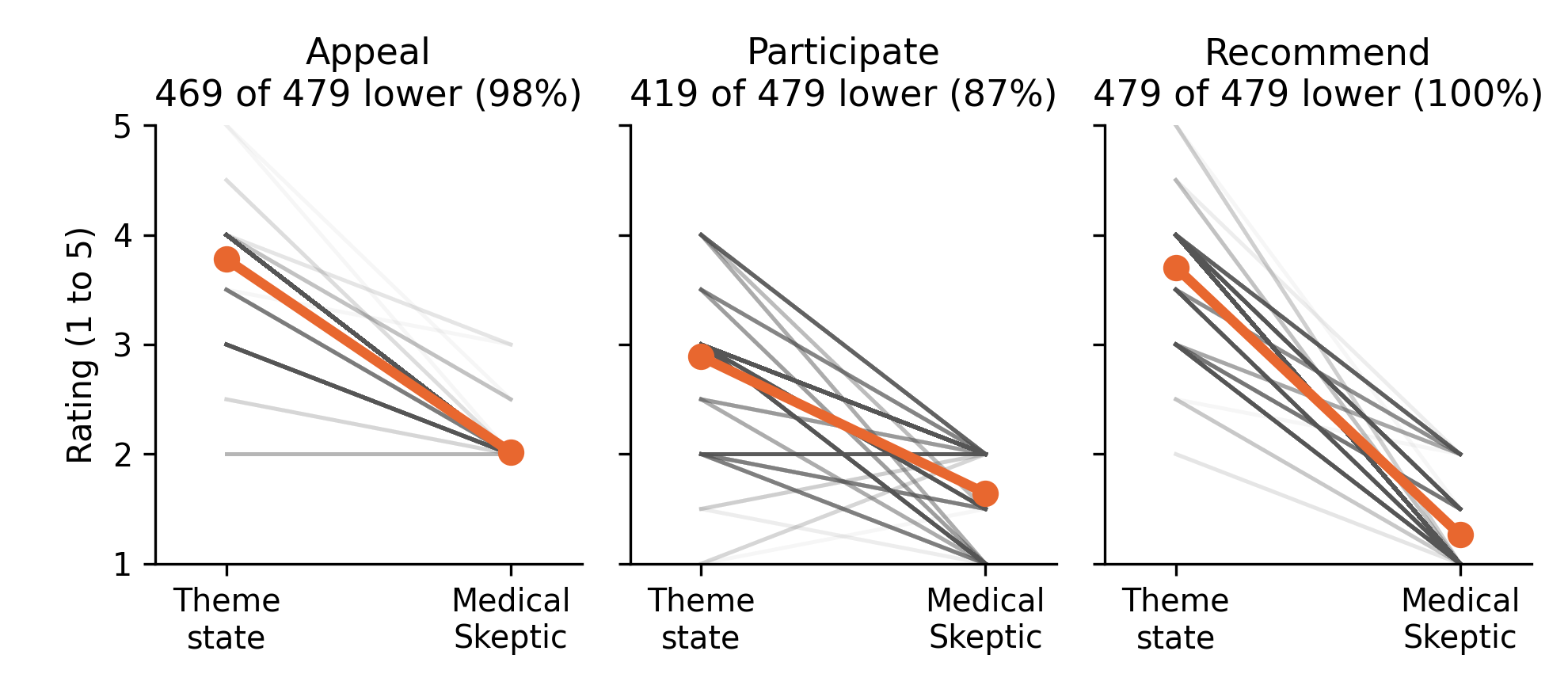}
  \caption{Within-person natural experiment: the 479 personas whose replicate samples landed, by chance, under both an active Medical Skeptic state and their theme's own state. The paired differences are large, consistent, and in the expected direction. Thin gray lines are individual personas (the mean of their replicates), and the orange line is the mean of all 479. Each panel title counts the personas whose rating is lower under the Medical Skeptic state than under their theme's own state.}
  \label{fig:within-person-slopes}
\end{figure}

\paragraph{Within-person check.} Figure~\ref{fig:within-person-slopes} shows the paired differences behind the natural-experiment result in \S\ref{sec:jmir-results}.

\paragraph{Why Arm B matches none.} Table~\ref{tab:jmir-cellmeans} shows the cause. In Arm B, the contribution-to-science theme at neutral trust has a mean appeal of 3.49, against 3.09 in Arm A. Every interaction is measured against that baseline cell, so its height sets the sign of all six coefficients. We report the pattern and have not tested why the theme sentence raises that cell.

\begin{table}[t]
  \centering
  \caption{Mean appeal rating (1 to 5) by arm, theme, and trust tier, with 604 to 805 personas in each cell. The three arms are generated for the same personas.}
  \label{tab:jmir-cellmeans}
  \small
  \begin{tabular}{llrrr}
    \toprule
    Arm & Theme & Distrust & Neutral & Trust \\
    \midrule
    A & Blood pressure          & 2.02 & 3.08 & 3.92 \\
    A & Community               & 2.02 & 3.16 & 4.00 \\
    A & Perks                   & 2.07 & 3.45 & 4.05 \\
    A & Contribution to science & 2.01 & 3.09 & 4.05 \\
    \midrule
    B & Blood pressure          & 2.15 & 3.17 & 3.95 \\
    B & Community               & 2.11 & 3.29 & 4.09 \\
    B & Perks                   & 2.27 & 3.61 & 4.02 \\
    B & Contribution to science & 2.04 & 3.49 & 4.10 \\
    \midrule
    C (domain-general) & Blood pressure          & 2.36 & 2.78 & 3.15 \\
    C (domain-general) & Community               & 2.43 & 3.12 & 3.56 \\
    C (domain-general) & Perks                   & 2.41 & 3.10 & 3.41 \\
    C (domain-general) & Contribution to science & 2.46 & 3.16 & 3.54 \\
    \bottomrule
  \end{tabular}
\end{table}

\subsection{Elicitation-Format Validation}
\label{sec:elicitation-format-validation}

Both empirical demonstrations use a hybrid \texttt{\{reasoning, rating\}} structured-output format, in which the model gives a short reasoning and then a rating. We chose it in a pilot on eight persona-context pairs from the motivational-themes study (20 samples each, Claude Haiku 4.5, temperature 0.7). The pairs were drawn from an earlier 20-pair pilot, five where a bare structured rating gave identical answers across all 20 samples and three where it did not. The pilot compared three formats, namely a bare structured rating, free text ending in a rating, and the hybrid. Table~\ref{tab:elicitation} gives the mean sampling variance of the appeal rating.

\begin{table}[h]
  \centering
  \small
  \caption{Mean sampling variance of the appeal rating over eight persona-context pairs, by elicitation format, with 95\% bootstrap intervals (2,000 resamples). Each format has 160 responses.}
  \label{tab:elicitation}
  \begin{tabular}{lccc}
    \toprule
    Format & Mean variance & 95\% interval & Parsed \\
    \midrule
    Bare structured rating & 0.090 & [0.025, 0.172] & 160 of 160 \\
    Free text ending in a rating & 0.289 & [0.093, 0.548] & 160 of 160 \\
    Hybrid (reasoning, then rating) & 0.164 & [0.098, 0.224] & 160 of 160 \\
    \bottomrule
  \end{tabular}
\end{table}

The bare format gave the lowest variance, and the hybrid recovered about 37\% of the gap between the bare format and free text. With eight pairs the intervals overlap, so this is a pilot-scale result. Low variance in the model's ratings would hide interaction effects whatever the activation mechanism does, so a format that does not suppress variance was a precondition for a fair test. We chose the hybrid over free text because its schema-validated output guarantees parseable responses at the scale of the main runs (about 25,000 responses for the motivational-themes study). Free text parsed at 100\% in this 160-response test, but one small run with one prompt template does not guarantee that at scale.

\section{Example Personas, Prompts, and Run Metadata}
\label{sec:prompts-examples}

This appendix shows what the language model receives in each arm, using the exact templates from our code. The example people are constructed with the same templates and are not real individuals. Wording that would identify the study's authors or institution is replaced by a bracketed note.

\subsection{Three Families, One Person and One Situation}
\label{sec:examples-families}

Table~\ref{tab:family-examples} shows the same constructed person facing the same ad in each family. The memory-evolving row is an illustration of the family and was not evaluated in our experiments.

\begin{table}[h]
  \centering
  \caption{What the language model receives about the person, by family. The situation (the ad) is the same in every row.}
  \label{tab:family-examples}
  \small
  \begin{tabular}{p{0.2\linewidth}p{0.72\linewidth}}
    \toprule
    Family & Person-level input to the model \\
    \midrule
    Fixed persona (Arm A) & ``Persona: Race: Black/African American; Age: 50-59; Sex: Female; Political view: Moderate; Location: Rural area; Household income: Less than \$35000; Education: Bachelor's degree; High blood pressure diagnosis: Yes. Based on your life experience, you generally distrust medical institutions and clinical research.'' \\
    Memory-evolving (illustration) & The same persona, plus a growing record of past events, for example ``Last month you read about a clinic that never shared results with participants.'' \\
    Person-in-context (Arm C) & The same demographics, with no trust sentence, plus an identity state selected for this situation: ``Right now, this person's salient concern is: You are wary of medical institutions and clinical research. You scrutinize who is running this, whether Black investigators are involved, and whether this could be another instance of institutions using people like you without giving back.'' \\
    \bottomrule
  \end{tabular}
\end{table}

\subsection{Motivational-Themes Study: Prompts by Arm}
\label{sec:examples-jmir}

All arms share the opening of the system prompt: ``You are simulating the survey response of a specific person for a research study on synthetic population simulation. Respond only as this person would. Do not break character or comment on the task.'' It is followed by ``Persona:'' and the demographic line from Table~\ref{tab:family-examples}.
\begin{itemize}[leftmargin=*]
  \item \emph{Arm A} adds the trust sentence for the person's tier. The three sentences are: ``Based on your life experience, you generally distrust medical institutions and clinical research.'' (distrust), ``You have a fairly neutral, wait-and-see attitude toward medical institutions and clinical research.'' (neutral), and ``Based on your life experience, you generally trust medical institutions and clinical research.'' (trust).
  \item \emph{Arm B} adds the same trust sentence plus one sentence naming the ad theme: ``This ad emphasizes contributing to scientific research.'' (research), ``This ad emphasizes improving the well-being of Black communities.'' (community), ``This ad emphasizes personal health benefit, specifically improving your own blood pressure.'' (blood pressure), or ``This ad emphasizes free groceries or cash compensation.'' (perks).
  \item \emph{Arm C} adds no trust sentence. It adds ``Right now, this person's salient concern is:'' followed by the norm of the selected identity state. The six states are Community \& Scientific Contribution, Health Focused, Medical Skeptic, Prudential Consumer, Privacy \& Surveillance Wary, and Faith-Centered Providence, and their full wording is in Appendix~\ref{sec:regime-spec-jmir}.
\end{itemize}
The user turn is identical in all arms and consists of the ad followed by the outcome question. The ad has the form ``[study name]; [theme headline]; A 12-week research study investigating the potential impact of fresh groceries on lowering blood pressure in Black adults. [contact line removed for review]''. The four headlines are ``Join our clinical trial to advance scientific research!'', ``Join our clinical trial to study a way to improve the well-being of Black communities!'', ``Join our clinical trial to study a dietary approach to improve your blood pressure!'', and ``Join our clinical trial to receive either free delivered fresh groceries or cash!'' The outcome question reads: ``Assume [the medical center] is a research hospital near where you live. Please review the clinical trial advertisement above and rate how much you agree or disagree with the following statements, each on a 5-point scale (5 = Strongly agree, 4 = Somewhat agree, 3 = Neither agree or disagree, 2 = Somewhat disagree, 1 = Strongly disagree): (1) appeal -- `This clinical trial recruitment advertisement is appealing to me'; (2) participate -- `I plan to call or email [the medical center] to participate in this clinical trial'; (3) recommend -- `I plan to share this clinical trial advertisement with others who may be eligible'. Give a brief reasoning, then the three ratings.'' The model returns a short reasoning and three integer ratings from 1 to 5 as structured output.

\subsection{Chameleon: Prompts by Arm}
\label{sec:examples-chameleon}

The persona line is built from the user's trait scores. We rank 21 trait dimensions (values, motivation, and risk attitudes) by how far they are from average, keep the five largest, and write one sentence for each in the high or low form. The five Big Five constructs are the outcome and are never in the persona. An example for a constructed user, whose top five traits include novelty-seeking and comfort with social risk, is: ``Persona: This person seeks novelty, excitement, and new experiences. This person is comfortable with social risks, like disagreeing publicly. This person isn't especially attached to tradition or custom. This person doesn't feel especially connected to other people right now. This person sticks closely to ethical rules.''

All arms share the opening: ``You are simulating how a specific person would come across in a piece of short writing, for a research study on synthetic population simulation. Respond only as this person would. Do not break character or comment on the task.'' The user turn ends with the same rating question in every arm: ``Rate how much each of the following describes how you're coming across right now, on a 1-5 scale (1 = not at all, 5 = very much): extraversion (outgoing, energetic), agreeableness (warm, cooperative), conscientiousness (organized, disciplined), neuroticism (anxious, emotionally reactive), openness (curious, open to new ideas). Give a brief reasoning, then the five ratings.''
\begin{itemize}[leftmargin=*]
  \item \emph{Arm A:} the user turn begins ``You are about to write a short post online.''
  \item \emph{Arm B:} the same persona, and the user turn begins ``You are about to write a short post in the r/[subreddit] community on Reddit.''
  \item \emph{Arm C:} the same user turn as Arm B, and the system prompt adds ``Right now, the aspect of this person's identity that's most active is:'' followed by up to two identity states from Appendix~\ref{sec:regime-spec-chameleon}, for example ``Primarily Expansive Engaged (socially confident and broadly engaged across many areas of life, comfortable taking risks in multiple domains).''
\end{itemize}

For Arm C, a separate call scores each subreddit once against the five activation cues. The system prompt reads ``You are scoring how well a Reddit community's typical content matches a set of behavioral/content cues, for a research study on context-activated identity. Score based on the community's general content and tone, not any specific post.'' The user turn reads ``Community: r/[subreddit]. For each set of cues below, score 0, 1, or 2 how well r/[subreddit]'s typical content matches it (0 = no match, 1 = some/background match, 2 = strong/central match)'', followed by the five cue descriptions, and the model returns a short reasoning and one score per state. The cue descriptions are, in the order of Table~\ref{tab:chameleon-wording}:
\begin{itemize}[leftmargin=*]
  \item \emph{Expansive Engaged.} Content involving active social engagement, competitive or achievement framing, novelty-seeking, or discussion of risk-taking across financial, recreational, or social domains, such as competitive gaming, entrepreneurship and investing, adventure or physical-risk activities, active dating, and high-energy outgoing social interaction.
  \item \emph{Amotivated Disinhibited.} Content reflecting disengagement, low investment, or venting without goal-orientation, combined with casual mentions of rule-bending, safety-cutting-corners, or gambling, such as low-effort complaining, confession-style rule-breaking, casual gambling or lottery discussion, and anonymous low-investment participation.
  \item \emph{High-Intensity Reactive.} Content involving high emotional intensity combined with high personal investment or striving, but not framed around risk-taking, such as high-stakes personal-crisis discussion, competitive achievement under pressure, and emotionally charged advice-seeking where the poster is highly engaged and anxious.
  \item \emph{Quiet Withdrawn.} Content that is low-stakes, passive, or solitary, such as casual browsing or lurker-friendly communities, simple hobby or leisure content without competitive or social-striving framing, and low posting or interaction demands.
  \item \emph{Conventional Risk-Tolerant.} Content that frames risk-taking within a conventional or practical, duty-oriented lens, such as personal finance or insurance framed around family responsibility, traditional-values-adjacent communities where practical rule-bending is normalized, and health or safety trade-offs made for practical or family reasons.
\end{itemize}

\subsection{Run Metadata}
\label{sec:run-metadata}

Table~\ref{tab:run-metadata} records what we have about the runs. We list what was not recorded, so that no one assumes it was.

\begin{table}[h]
  \centering
  \caption{Run metadata. Costs are estimates from token counts at \$1 and \$5 per million input and output tokens, and are not billed amounts.}
  \label{tab:run-metadata}
  \small
  \begin{tabular}{p{0.3\linewidth}p{0.62\linewidth}}
    \toprule
    Item & Value \\
    \midrule
    Model & Claude Haiku 4.5 (model alias \texttt{claude-haiku-4-5}; no dated snapshot recorded) \\
    Sampling & Temperature 0.7. Output requested as a JSON schema. Up to 3 attempts, retried only when the output could not be parsed. \\
    Output length limit & 300 tokens (motivational-themes study), 400 tokens (Chameleon full run), 250 tokens (Chameleon pilot) \\
    Run dates & 12 to 13 August 2026 \\
    Chameleon full run & 31,352 responses: 10,452 for Arm A, 10,448 for Arm B, 10,452 for Arm C. Two replicates for 1,517 users and three for 150 users. \\
    Motivational-themes study & 24,870 responses (8,290 per arm) for $n=6{,}632$ personas, with three replicates for the first 829 personas and one for the rest. The theme-amplified refinement adds 8,290 responses. \\
    Total logged & About 71,000 model calls, at an estimated cost of about \$96 \\
    Random seeds & Persona sampling 20260812. Mixture model: \texttt{random\_state} 0, 50 initializations, tied covariance, 6 components. Bootstrap: 1,000 resamples, seed 0. The model's own sampling cannot be seeded. \\
    Not recorded & Counts of failed or unparseable responses, retry counts, request identifiers, and the software versions used at run time \\
    \bottomrule
  \end{tabular}
\end{table}

%% file: figures/fig_appendix_chameleon_pipeline.tex
\begin{tikzpicture}[x=1cm,y=1cm,font=\normalsize,
  box/.style={draw=black!60,rounded corners=2pt,align=left,inner sep=4pt,text width=14.6cm},
  half/.style={draw=black!60,rounded corners=2pt,align=left,inner sep=4pt,text width=7.0cm},
  data/.style={fill=blue!8},  calc/.style={fill=green!8},  judge/.style={fill=orange!14},
  subj/.style={fill=red!10},  real/.style={fill=violet!9},
  tag/.style={font=\footnotesize\bfseries,text=white,rounded corners=1pt,inner sep=2pt,anchor=south east},
  arr/.style={-{Stealth[length=4pt]},thick,black!70}]
\node[box,data,anchor=north] (n1) at (0,0) {\textbf{1. Real data.} 1,667 Reddit users, three posts each. Each user has 26 trait scores (5 Big Five, 10 values, 5 motivation, 6 risk), standardized and averaged over the posts.};
\node[box,calc,anchor=north] (n2) at ([yshift=-0.5cm]n1.south) {\textbf{2. Group the users.} A Gaussian mixture model fits 6 groups. Each user gets a soft membership across the groups, which is the user's identity-state profile.};
\node[half,judge,anchor=north west] (n3a) at ([yshift=-0.5cm]n2.south west) {\textbf{3a. Name the groups (once).} Three blind model sessions see only each group's average trait pattern. Each proposes a name, literature support, and activation cues. The largest, most average group stays as the baseline and is not an activated state.};
\node[half,calc,anchor=north east] (n3b) at ([yshift=-0.5cm]n2.south east) {\textbf{3b. Affinity (per person).} The weight for each state is 1.0, plus a bounded boost for the person's top two states (0.5 and 0.25). Raw membership was too extreme: for 95.9\% of users one state decided everything, so context could not compete.};
\node[half,judge,anchor=north east] (n4) at ([yshift=-0.5cm]n3b.south east) {\textbf{4. Context cues.} For each subreddit, a model scores how well its usual content matches each state's activation cues: 0, 1, or 2.};
\path let \p1=(n3a.south), \p2=(n4.south), \n1={min(\y1,\y2)-0.5cm} in coordinate (rowC) at (0,\n1);
\node[box,calc,anchor=north] (n5) at (rowC) {\textbf{5. Active state for each (person, subreddit).} The state with the highest affinity times cue score is active. The baseline is used if no cue matches.};
\node[box,subj,anchor=north] (n6) at ([yshift=-0.5cm]n5.south) {\textbf{6. Generate behavior.} The language model acts out the person. Arm A: profile only. Arm B: profile plus the subreddit. Arm C: profile plus the active state.};
\node[box,real,anchor=north] (n7) at ([yshift=-0.5cm]n6.south) {\textbf{7. Real target, no model involved.} Each person's real shift in each context, split into the state-typical part and the person-specific remainder.};
\node[box,subj,anchor=north] (n8) at ([yshift=-0.5cm]n7.south) {\textbf{8. Score each arm against the real data.} Between-person share (ICC) against the real value, and correlation of the simulated signal with the real person-specific signal.};
\node[tag,fill=orange!80!black] at (n3a.north east) {model = judge};
\node[tag,fill=orange!80!black] at (n4.north east) {model = judge};
\node[tag,fill=red!70!black] at (n6.north east) {model = simulated person};
\draw[arr] (n1.south) -- (n2.north);
\draw[arr] (n2.south -| n3a) -- (n3a.north);
\draw[arr] (n2.south -| n3b) -- (n3b.north);
\draw[arr] (n3b.south) -- (n4.north);
\draw[arr] (n3a.south) -- (n3a.south |- n5.north);
\draw[arr] (n4.south) -- (n4.south |- n5.north);
\draw[arr] (n5.south) -- (n6.north);
\draw[arr] (n6.south) -- (n7.north);
\draw[arr] (n7.south) -- (n8.north);
\end{tikzpicture}

%% file: figures/fig_appendix_jmir_pipeline.tex
\begin{tikzpicture}[x=1cm,y=1cm,font=\normalsize,
  box/.style={draw=black!60,rounded corners=2pt,align=left,inner sep=4pt,text width=14.6cm},
  half/.style={draw=black!60,rounded corners=2pt,align=left,inner sep=4pt,text width=7.0cm},
  data/.style={fill=blue!8},  calc/.style={fill=green!8},  judge/.style={fill=orange!14},
  subj/.style={fill=red!10},  gate/.style={fill=yellow!14},
  tag/.style={font=\footnotesize\bfseries,text=white,rounded corners=1pt,inner sep=2pt,anchor=south east},
  arr/.style={-{Stealth[length=4pt]},thick,black!70}]
\node[box,data,anchor=north] (n1) at (0,0) {\textbf{1. Reproduction target.} A published study of 829 Black adults, each shown one of four ad themes. Targets: no theme main effect, and six blood-pressure by trust coefficients that are negative at both trust extremes.};
\node[box,calc,anchor=north] (n2) at ([yshift=-0.5cm]n1.south) {\textbf{2. Write six identity states first,} from the health-behavior and clinical-recruitment literature and independent of the target study's results (the reverse order from Chameleon). The risk of prior exposure to the target study is checked in step 3.};
\node[box,judge,anchor=north] (n3) at ([yshift=-0.5cm]n2.south) {\textbf{3. Circularity check.} Three blind model sessions with no access to the specification or the target study's results. They merged two states, added one (Faith-Centered Providence), and agreed most on Medical Skeptic.};
\node[half,calc,anchor=north west] (n4a) at ([yshift=-0.5cm]n3.south west) {\textbf{4a. Affinity (per person).} 1.0, plus literature-based demographic modifiers (income, age, education) and small per-person noise.};
\node[half,calc,anchor=north east] (n4b) at ([yshift=-0.5cm]n3.south east) {\textbf{4b. Context cues.} Fixed scores of 0, 1, or 2 for how well each ad theme matches each state's cues, set in advance.};
\path let \p1=(n4a.south), \p2=(n4b.south), \p3=(n4a.west), \p4=(n4b.east), \n1={min(\y1,\y2)-0.5cm} in coordinate (rowL) at (\x3,\n1) coordinate (rowR) at (\x4,\n1);
\node[half,gate,anchor=north west] (n5a) at (rowL) {\textbf{5a. Medical Skeptic: direct trust gate.} Checked first for each (person, theme). Activation probability is a base set by trust (distrust 0.70, neutral 0.35, trust 0.15), plus a floor of 0.05 and 0.03 per demographic match, capped. It does not compete with the other states.};
\node[half,calc,anchor=north east] (n5b) at (rowR) {\textbf{5b. The other five states.} A weighted competition of affinity times cue score, and the top choice wins when the gate does not fire. Two of the five never activate for these ads, because no ad contains their cues. This is by design.};
\path let \p1=(n5a.south), \p2=(n5b.south), \n1={min(\y1,\y2)-0.5cm} in coordinate (rowM) at (0,\n1);
\node[box,calc,anchor=north] (n6) at (rowM) {\textbf{6. Active state for each (person, ad theme),} from whichever path fires: the trust-gate draw or the competition among the other five.};
\node[box,subj,anchor=north] (n7) at ([yshift=-0.5cm]n6.south) {\textbf{7. Generate behavior.} The language model acts out the person. Arm A: demographics plus a fixed trust sentence. Arm B: Arm A plus one sentence naming the theme. Arm C: demographics plus the active state, with no trust sentence.};
\node[box,data,anchor=north] (n8) at ([yshift=-0.5cm]n7.south) {\textbf{8. Compare with the real target.} There is one observation per real person, so no repeated-measures test is possible. (1) Trust main effect: does simulated appeal rise from distrust to trust? (2) Direction of the six blood-pressure by trust coefficients (Table~\ref{tab:jmir-specs}).};
\node[tag,fill=orange!80!black] at (n3.north east) {model = judge};
\node[tag,fill=red!70!black] at (n7.north east) {model = simulated person};
\draw[arr] (n1.south) -- (n2.north);
\draw[arr] (n2.south) -- (n3.north);
\draw[arr] (n3.south -| n4a) -- (n4a.north);
\draw[arr] (n3.south -| n4b) -- (n4b.north);
\draw[arr] (n4a.south) -- (n5a.north);
\draw[arr] (n4b.south) -- (n5b.north);
\draw[arr] (n5a.south) -- (n5a.south |- n6.north);
\draw[arr] (n5b.south) -- (n5b.south |- n6.north);
\draw[arr] (n6.south) -- (n7.north);
\draw[arr] (n7.south) -- (n8.north);
\end{tikzpicture}

%% file: references.bib
@inproceedings{zheng2023judging,
  author    = {Zheng, Lianmin and Chiang, Wei-Lin and Sheng, Ying and Zhuang, Siyuan and Wu, Zhanghao and Zhuang, Yonghao and Lin, Zi and Li, Zhuohan and Li, Dacheng and Xing, Eric P. and Zhang, Hao and Gonzalez, Joseph E. and Stoica, Ion},
  title     = {Judging {LLM}-as-a-Judge with {MT}-Bench and Chatbot Arena},
  booktitle = {Proceedings of the 37th International Conference on Neural Information Processing Systems (NIPS '23)},
  year      = {2023},
  pages     = {Article 2020, 46595--46623},
  publisher = {Curran Associates Inc.},
  address   = {Red Hook, NY, USA},
}

@article{gu2024survey,
  author  = {Gu, Jiawei and Jiang, Xuhui and Shi, Zhichao and Tan, Hexiang and Zhai, Xuehao and Xu, Chengjin and Li, Wei and Shen, Yinghan and Ma, Shengjie and Liu, Honghao and Wang, Saizhuo and Zhang, Kun and Lin, Zhouchi and Zhang, Bowen and Ni, Lionel and Gao, Wen and Wang, Yuanzhuo and Guo, Jian},
  title   = {A Survey on {LLM}-as-a-Judge},
  journal = {The Innovation},
  year    = {2026},
  volume  = {7},
  number  = {6},
  pages   = {101253},
  doi     = {10.1016/j.xinn.2025.101253},
}

@inproceedings{li2026preferenceleakage,
  author    = {Li, Dawei and Sun, Renliang and Huang, Yue and Zhong, Ming and Jiang, Bohan and Han, Jiawei and Zhang, Xiangliang and Wang, Wei and Liu, Huan},
  title     = {Preference Leakage: A Contamination Problem in {LLM}-as-a-Judge},
  booktitle = {Proceedings of the Fourteenth International Conference on Learning Representations (ICLR 2026)},
  year      = {2026},
}

@article{scharff2010more,
  author  = {Scharff, Darcell P. and Mathews, Katherine J. and Jackson, Pamela and Hoffsuemmer, Jonathan and Martin, Emeobong and Edwards, Dorothy},
  title   = {More than {Tuskegee}: Understanding Mistrust about Research Participation},
  journal = {Journal of Health Care for the Poor and Underserved},
  year    = {2010},
  volume  = {21},
  number  = {3},
  pages   = {879--897},
  doi     = {10.1353/hpu.0.0323},
}

@article{armstrong2013prior,
  author  = {Armstrong, Katrina and Putt, Mary and Halbert, Chanita Hughes and Grande, David and Schwartz, J. Sanford and Liao, Kaijun and Marcus, Noora and Demeter, Mirar Bristol and Shea, Judy A.},
  title   = {Prior Experiences of Racial Discrimination and Racial Differences in Health Care System Distrust},
  journal = {Medical Care},
  year    = {2013},
  volume  = {51},
  number  = {2},
  pages   = {144--150},
  doi     = {10.1097/MLR.0b013e31827310a1},
}

@article{alsan2018tuskegee,
  author  = {Alsan, Marcella and Wanamaker, Marianne},
  title   = {{Tuskegee} and the Health of Black Men},
  journal = {Quarterly Journal of Economics},
  year    = {2018},
  volume  = {133},
  number  = {1},
  pages   = {407--455},
  doi     = {10.1093/qje/qjx029},
}

@article{wu2026humanlm,
  author  = {Wu, Shirley and Choi, Evelyn and Khatua, Arpandeep and Wang, Zhanghan and He-Yueya, Joy and Weerasooriya, Tharindu Cyril and Wei, Wei and Yang, Diyi and Leskovec, Jure and Zou, James},
  title   = {{HumanLM}: Simulating Users with State Alignment Beats Response Imitation},
  journal = {arXiv preprint arXiv:2603.03303},
  year    = {2026},}

@article{harry2026jmir,
  author  = {Harry, Tamunotonye and others},
  title   = {Guide on Selection of Optimal Motivational Themes for Use in a Clinical Trial Recruiting Black {US} Adults: Survey Study},
  journal = {Journal of Medical Internet Research},
  year    = {2026},
  volume  = {28},
  pages   = {e75857},
  doi     = {10.2196/75857},
  url     = {https://www.jmir.org/2026/1/e75857/},
}

@inproceedings{harry2026acl,
  author    = {Harry, Tamunotonye and Ngong, Ivoline C. and Nweke, Chima and Feng, Yuanyuan and Near, Joseph},
  title     = {Beyond Fixed Psychological Personas: State Beats Trait, but Language Models are State-Blind},
  booktitle = {Findings of the Association for Computational Linguistics: ACL 2026},
  year      = {2026},
  address   = {San Diego, California, United States},
  publisher = {Association for Computational Linguistics},
  pages     = {26440--26468},
  doi       = {10.18653/v1/2026.findings-acl.1316},
  note = {arXiv:2601.15395},
}

@article{akerlofkranton2000,
  author    = {Akerlof, George A. and Kranton, Rachel E.},
  title     = {Economics and Identity},
  journal   = {Quarterly Journal of Economics},
  year      = {2000},
  volume    = {115},
  number    = {3},
  pages     = {715--753},
  doi       = {10.1162/003355300554881},
}

@book{akerlofkranton2010,
  author    = {Akerlof, George A. and Kranton, Rachel E.},
  title     = {Identity Economics: How Our Identities Shape Our Work, Wages, and Well-Being},
  publisher = {Princeton University Press},
  address   = {Princeton, NJ},
  year      = {2010},
}

@article{tettguterman2000,
  author  = {Tett, Robert P. and Guterman, Hal A.},
  title   = {Situation Trait Relevance, Trait Expression, and Cross-Situational Consistency: Testing a Principle of Trait Activation},
  journal = {Journal of Research in Personality},
  year    = {2000},
  volume  = {34},
  number  = {4},
  pages   = {397--423},
  doi     = {10.1006/jrpe.2000.2292},
}

@article{fleeson2001toward,
  author  = {Fleeson, William},
  title   = {Toward a Structure- and Process-Integrated View of Personality: Traits as Density Distributions of States},
  journal = {Journal of Personality and Social Psychology},
  year    = {2001},
  volume  = {80},
  number  = {6},
  pages   = {1011--1027},
  doi     = {10.1037/0022-3514.80.6.1011},
}

@article{bisbee2024synthetic,
  author  = {Bisbee, James and Clinton, Joshua D. and Dorff, Cassy and Kenkel, Brenton and Larson, Jennifer M.},
  title   = {Synthetic Replacements for Human Survey Data? The Perils of Large Language Models},
  journal = {Political Analysis},
  year    = {2024},
  volume  = {32},
  number  = {4},
  pages   = {401--416},
  doi     = {10.1017/pan.2024.5},
}

@inproceedings{sun2025sociodemographic,
  author    = {Sun, Huaman and Pei, Jiaxin and Choi, Minje and Jurgens, David},
  title     = {Sociodemographic Prompting is Not Yet an Effective Approach for Simulating Subjective Judgments with {LLM}s},
  booktitle = {Proceedings of the 2025 Conference of the Nations of the Americas Chapter of the Association for Computational Linguistics: Human Language Technologies (Volume 2: Short Papers)},
  year      = {2025},
  pages     = {845--854},
  publisher = {Association for Computational Linguistics},
  address   = {Albuquerque, New Mexico},
  doi       = {10.18653/v1/2025.naacl-short.71},
}

@misc{li2026matraix,
  author = {Li, Xiaomin and Hao, Yuexing and others},
  title  = {MatrAIx: Simulating the World with 8.3 Billion Persona Agents},
  year   = {2026},
  note = {arXiv:2608.04205},
}

@misc{park2024interviews,
  author = {Park, Joon Sung and Zou, Carolyn Q. and Kamphorst, Jonne and Egan, Niles and Shaw, Aaron and Hill, Benjamin Mako and Cai, Carrie and Morris, Meredith Ringel and Liang, Percy and Willer, Robb and Bernstein, Michael S.},
  title  = {{LLM} Agents Grounded in Self-Reports Enable General-Purpose Simulation of Individuals},
  year   = {2024},
  note = {arXiv:2411.10109},
}

@inproceedings{park2023generativeagents,
  author    = {Park, Joon Sung and O'Brien, Joseph C. and Cai, Carrie J. and Morris, Meredith Ringel and Liang, Percy and Bernstein, Michael S.},
  title     = {Generative Agents: Interactive Simulacra of Human Behavior},
  booktitle = {Proceedings of the 36th Annual ACM Symposium on User Interface Software and Technology (UIST '23)},
  year      = {2023},
  publisher = {Association for Computing Machinery},
  address   = {New York, NY, USA},
  doi       = {10.1145/3586183.3606763},
}

@article{argyle2023demographic,
  author  = {Argyle, Lisa P. and Busby, Ethan C. and Fulda, Nancy and Gubler, Joshua and Rytting, Christopher and Wingate, David},
  title   = {Out of One, Many: Using Language Models to Simulate Human Samples},
  journal = {Political Analysis},
  year    = {2023},
  volume  = {31},
  number  = {3},
  pages   = {337--351},
  doi     = {10.1017/pan.2023.2},
}

@inproceedings{zhou2024sotopia,
  author    = {Zhou, Xuhui and Zhu, Hao and Mathur, Leena and Zhang, Ruohong and Yu, Haofei and Qi, Zhengyang and Morency, Louis-Philippe and Bisk, Yonatan and Fried, Daniel and Neubig, Graham and Sap, Maarten},
  title     = {{SOTOPIA}: Interactive Evaluation for Social Intelligence in Language Agents},
  booktitle = {Proceedings of the Twelfth International Conference on Learning Representations (ICLR 2024)},
  year      = {2024},
  note = {arXiv:2310.11667},
}

@misc{vezhnevets2023concordia,
  author = {Vezhnevets, Alexander Sasha and Agapiou, John P. and Aharon, Avia and Ziv, Ron and Matyas, Jayd and Du{\'e}{\~n}ez-Guzm{\'a}n, Edgar A. and Cunningham, William A. and Osindero, Simon and Karmon, Danny and Leibo, Joel Z.},
  title  = {Generative Agent-Based Modeling with Actions Grounded in Physical, Social, or Digital Space Using Concordia},
  year   = {2023},
  note = {arXiv:2312.03664},
}

@misc{yang2024oasis,
  author = {Yang, Ziyi and Zhang, Zaibin and Zheng, Zirui and Jiang, Yuxian and Gan, Ziyue and Wang, Zhiyu and Ling, Zijian and Chen, Jinsong and Ma, Martz and Dong, Bowen and Gupta, Prateek and Hu, Shuyue and Yin, Zhenfei and Li, Guohao and Jia, Xu and Wang, Lijun and Ghanem, Bernard and Lu, Huchuan and Lu, Chaochao and Ouyang, Wanli and Qiao, Yu and Torr, Philip and Shao, Jing},
  title  = {{OASIS}: Open Agent Social Interaction Simulations with One Million Agents},
  year   = {2024},
  note = {arXiv:2411.11581},
}

@misc{piao2025agentsociety,
  author = {Piao, Jinghua and Yan, Yuwei and Zhang, Jun and Li, Nian and Yan, Junbo and Lan, Xiaochong and Lu, Zhihong and Zheng, Zhiheng and Wang, Jing Yi and Zhou, Di and Gao, Chen and Xu, Fengli and Zhang, Fang and Rong, Ke and Su, Jun and Li, Yong},
  title  = {{AgentSociety}: Large-Scale Simulation of {LLM}-Driven Generative Agents Advances Understanding of Human Behaviors and Society},
  year   = {2025},
  note = {arXiv:2502.08691},
}

@article{fleesonjayawickreme2015,
  author  = {Fleeson, William and Jayawickreme, Eranda},
  title   = {Whole Trait Theory},
  journal = {Journal of Research in Personality},
  year    = {2015},
  volume  = {56},
  pages   = {82--92},
  doi     = {10.1016/j.jrp.2014.10.009},
}

@article{steyer1992laten,
  author  = {Steyer, Rolf and Ferring, Dieter and Schmitt, Manfred J.},
  title   = {States and Traits in Psychological Assessment},
  journal = {European Journal of Psychological Assessment},
  year    = {1992},
  volume  = {8},
  number  = {2},
  pages   = {79--98},
}

@article{steyer2015laten,
  author  = {Steyer, Rolf and Mayer, Axel and Geiser, Christian and Cole, David A.},
  title   = {A Theory of States and Traits---Revised},
  journal = {Annual Review of Clinical Psychology},
  year    = {2015},
  volume  = {11},
  pages   = {71--98},
  doi     = {10.1146/annurev-clinpsy-032813-153719},
}

@article{eiddiener2004,
  author  = {Eid, Michael and Diener, Ed},
  title   = {Global Judgments of Subjective Well-Being: Situational Variability and Long-Term Stability},
  journal = {Social Indicators Research},
  year    = {2004},
  volume  = {65},
  number  = {3},
  pages   = {245--277},
}

@book{turner1987selfcategorization,
  author    = {Turner, John C. and Hogg, Michael A. and Oakes, Penelope J. and Reicher, Stephen D. and Wetherell, Margaret S.},
  title     = {Rediscovering the Social Group: A Self-Categorization Theory},
  publisher = {Basil Blackwell},
  address   = {Oxford},
  year      = {1987},
}

@article{mcconnell2011multiple,
  author  = {McConnell, Allen R.},
  title   = {The Multiple Self-Aspects Framework: Self-Concept Representation and Its Implications},
  journal = {Personality and Social Psychology Review},
  year    = {2011},
  volume  = {15},
  number  = {1},
  pages   = {3--27},
  doi     = {10.1177/1088868310371101},
}

@book{raudenbushbryk2002,
  author    = {Raudenbush, Stephen W. and Bryk, Anthony S.},
  title     = {Hierarchical Linear Models: Applications and Data Analysis Methods},
  edition   = {2nd},
  publisher = {Sage Publications},
  address   = {Thousand Oaks, CA},
  year      = {2002},
}

@book{snijdersbosker2011,
  author    = {Snijders, Tom A. B. and Bosker, Roel J.},
  title     = {Multilevel Analysis: An Introduction to Basic and Advanced Multilevel Modeling},
  edition   = {2nd},
  publisher = {Sage Publications},
  address   = {London},
  year      = {2011},
}

@article{wang2025roleplay,
  author  = {Wang, Angelina and Morgenstern, Jamie and Dickerson, John P.},
  title   = {Large Language Models that Replace Human Participants can Harmfully Misportray and Flatten Identity Groups},
  journal = {Nature Machine Intelligence},
  year    = {2025},
  volume  = {7},
  number  = {3},
  pages   = {400--411},
  doi     = {10.1038/s42256-025-00986-z},
}

@misc{ozkan2026distribution,
  author = {Ozkan, Gurkan},
  title  = {Distribution-First Population Simulation: Collapse, Calibration, and Recall in Non-{WEIRD} {LLM} Persona Modeling},
  year   = {2026},
  note = {arXiv:2607.18310},
}

@inproceedings{gonnermann2026persona,
  author    = {Gonnermann-M{\"u}ller, Jana and Haase, Jennifer and Leins, Nicolas and Kosch, Thomas and Pokutta, Sebastian},
  title     = {Maintaining Stable Personas? Examining Temporal Stability in {LLM}-Based Human Simulation},
  booktitle = {Extended Abstracts of the 2026 CHI Conference on Human Factors in Computing Systems (CHI EA '26)},
  year      = {2026},
  doi       = {10.1145/3772363.3799334},}

@inproceedings{morocho2026multiattribute,
  author    = {Taday Morocho, Erika Elizabeth and Cima, L. and Fagni, T. and Avvenuti, M. and Cresci, S.},
  title     = {Assessing the Reliability of Persona-Conditioned {LLM}s as Synthetic Survey Respondents},
  booktitle = {Companion Proceedings of the ACM Web Conference 2026 (WWW '26 Companion)},
  year      = {2026},
  pages     = {320--329},
  publisher = {Association for Computing Machinery},
  address   = {New York, NY, USA},
  doi       = {10.1145/3774905.3795477},
}

@misc{xiao2026collapse,
  author = {Xiao, Y. and Zhang, V. J. and Yang, C. and Ma, N. and Xuan, W. and Huang, J.},
  title  = {The Chameleon's Limit: Investigating Persona Collapse and Homogenization in Large Language Models},
  year   = {2026},
  note = {arXiv:2604.24698},
}

@inproceedings{bender2021parrots,
  author    = {Bender, Emily M. and Gebru, Timnit and McMillan-Major, Angelina and Shmitchell, Shmargaret},
  title     = {On the Dangers of Stochastic Parrots: Can Language Models Be Too Big?},
  booktitle = {Proceedings of the 2021 ACM Conference on Fairness, Accountability, and Transparency},
  year      = {2021},
  pages     = {610--623},
  publisher = {Association for Computing Machinery},
  address   = {New York, NY, USA},
  doi       = {10.1145/3442188.3445922},
}

@inproceedings{blodgett2020language,
  author    = {Blodgett, Su Lin and Barocas, Solon and Daum{\'e} III, Hal and Wallach, Hanna},
  title     = {Language (Technology) is Power: A Critical Survey of ``Bias'' in {NLP}},
  booktitle = {Proceedings of the 58th Annual Meeting of the Association for Computational Linguistics},
  year      = {2020},
  pages     = {5454--5476},
  publisher = {Association for Computational Linguistics},
  address   = {Online},
  doi     = {10.18653/v1/2020.acl-main.485},
}

@inproceedings{batzner2025representativeness,
  author    = {Batzner, Jan and Stocker, Volker and Tang, Bingjun and Natarajan, Anusha and Chen, Qinhao and Schmid, Stefan and Kasneci, Gjergji},
  title     = {Whose Personae? Synthetic Persona Experiments in {LLM} Research and Pathways to Transparency},
  booktitle = {Proceedings of the AAAI/ACM Conference on AI, Ethics, and Society},
  year      = {2025},
  volume    = {8},
  number    = {1},
  pages     = {343--354},
  doi       = {10.1609/aies.v8i1.36553},
}

@incollection{pielage2026dynamic,
  author    = {Pielage, Leon and H{\"a}tscher, Ole and Back, Mitja and Marschall, Bernhard and Risse, Benjamin},
  title     = {Dynamic Personality Adaptation in Large Language Models via State Machines},
  booktitle = {Pattern Recognition: 28th International Conference, {ICPR} 2026, Lyon, France, August 17--22, 2026, Proceedings, Part {VI}},
  editor    = {De Marsico, Maria and Ho, Tin Kam and Jurie, Frederic and Liu, Cheng-Lin and Lopresti, Daniel and Nystr{\"o}m, Ingela and Ogier, Jean-Marc and Ross, Arun and Wang, Liang},
  series    = {Lecture Notes in Computer Science},
  year      = {2026},
  pages     = {509--523},
  publisher = {Springer Nature Switzerland},
  address   = {Cham},
  doi       = {10.1007/978-3-032-31673-8_34},
}

@inproceedings{wei2026situational,
  author    = {Wei, Zesheng and Li, Mengxiang and Wang, Zilei and Deng, Yang},
  title     = {Beyond Static Personas: Situational Personality Steering for Large Language Models},
  booktitle = {Findings of the Association for Computational Linguistics: ACL 2026},
  year      = {2026},
  pages     = {19185--19210},
  publisher = {Association for Computational Linguistics},
  address   = {San Diego, California, United States},
  doi       = {10.18653/v1/2026.findings-acl.958},
}

@inproceedings{lee2025spectrum,
  author    = {Lee, Keyeun and Kim, Seo Hyeong and Lee, Seolhee and Eun, Jinsu and Ko, Yena and Jeon, Hayeon and Kim, Esther Hehsun and Cho, Seonghye and Yang, Soeun and Kim, Eun-mee and Lim, Hajin},
  title     = {{SPeCtrum}: A Grounded Framework for Multidimensional Identity Representation in {LLM}-Based Agent},
  booktitle = {Proceedings of the 2025 Conference of the Nations of the Americas Chapter of the Association for Computational Linguistics: Human Language Technologies (Volume 1: Long Papers)},
  year      = {2025},
  pages     = {6971--6991},
  publisher = {Association for Computational Linguistics},
  address   = {Albuquerque, New Mexico},
  doi       = {10.18653/v1/2025.naacl-long.356},
}

@misc{wang2026lifemem,
  author = {Wang, Hexi and Zhou, Yujia and Du, Bangde and Su, Weihang and Cao, Xinyuan and Pan, Qingyi and Ai, Qingyao and Wu, Yueyue and Zhang, Min and Liu, Yiqun},
  title = {Mitigating Identity Essentialism in {LLM} Agents with Longitudinal Life Trajectories},
  year = {2026}, eprint = {2608.19621}, archivePrefix = {arXiv}}

@misc{han2026personality,
  author = {Han, Bin and Kwon, Deuksin and Gratch, Jonathan},
  title = {Personality Expression Across Contexts: Linguistic and Behavioral Variation in {LLM} Agents},
  year = {2026}, eprint = {2602.01063}, archivePrefix = {arXiv}}

@article{fleeson2009implications,
  author  = {Fleeson, William and Gallagher, Patrick},
  title   = {The Implications of {Big-Five} Standing for the Distribution of Trait Manifestation in Behavior: Fifteen Experience-Sampling Studies and a Meta-Analysis},
  journal = {Journal of Personality and Social Psychology},
  year    = {2009},
  volume  = {97},
  number  = {6},
  pages   = {1097--1114},
  doi     = {10.1037/a0016786},
}
